%% file: main.tex
\documentclass[conference]{IEEEtran}
\IEEEoverridecommandlockouts

\usepackage{cite}
\usepackage{amsmath,amssymb,amsfonts}
\usepackage{algorithmic}
\usepackage{graphicx}
\usepackage{textcomp}
\usepackage{xcolor}

\usepackage{algorithm}
\usepackage{algorithmic}

\usepackage{amsthm}
\usepackage{amsmath}
\usepackage{amssymb}
\newtheorem{definition}{Definition}
\newtheorem{remark}{Remark}
\usepackage{cleveref}
\usepackage{booktabs}
\usepackage{multicol}
\def\BibTeX{{\rm B\kern-.05em{\sc i\kern-.025em b}\kern-.08em
    T\kern-.1667em\lower.7ex\hbox{E}\kern-.125emX}}
\renewcommand{\IEEEauthorrefmark}[1]{\textsuperscript{#1}}

\begin{document}

\title{Differentially Private Adaptation of Diffusion Models via Noisy Aggregated Embeddings}

\author{\IEEEauthorblockN{Pura Peetathawatchai\IEEEauthorrefmark{1,2}$^\dagger$\thanks{$^{\dagger}$Work done while at Stanford University.}\thanks{This work has been accepted for publication at the IEEE Conference on Secure and Trustworthy Machine Learning (SaTML 2026). The final version is/will be available on IEEE Xplore.}, Wei-Ning Chen\IEEEauthorrefmark{3}, Berivan Isik\IEEEauthorrefmark{4}, Sanmi Koyejo\IEEEauthorrefmark{1} and Albert No\IEEEauthorrefmark{5}}
\IEEEauthorblockA{\IEEEauthorrefmark{1}Stanford University, 
\IEEEauthorrefmark{2}ETH Zurich,
\IEEEauthorrefmark{3}Microsoft, 
\IEEEauthorrefmark{4}Google DeepMind, 
\IEEEauthorrefmark{5}Yonsei University\\
albertno@yonsei.ac.kr}}

\newcommand{\eve}{{\texttt{@eveismyname}}}
\newcommand{\clip}{\mathcal{E}}
\newcommand{\Dsub}{D_{\text{sub}}}

\newcommand{\lp}{\left(}
\newcommand{\rp}{\right)}
\newcommand{\lb}{\left[}
\newcommand{\rb}{\right]}
\newcommand{\lbp}{\left\{}
\newcommand{\rbp}{\right\}}
\newcommand{\lba}{\left\lvert}
\newcommand{\rba}{\right\rvert}
\newcommand{\lV}{\left\lVert}
\newcommand{\rV}{\right\rVert}
\newcommand{\mv}{\middle\vert}
\newcommand{\ul}{\underline}
\newcommand{\ol}{\overline}
\newcommand{\mcal}{\mathcal}
\newcommand{\mscr}{\mathscr}
\newcommand{\what}{\widehat}
\newcommand{\wtild}{\widetilde}
\newcommand{\mb}{\mathbf}
\newcommand{\bbm}{\mathbbm}
\newcommand{\mbb}{\mathbb}
\newcommand{\msf}{\mathsf}
\newcommand{\la}{\leftarrow}
\newcommand{\ra}{\rightarrow}
\newcommand{\ua}{\uparrow}
\newcommand{\da}{\downarrow}
\newcommand{\lra}{\leftrightarrow}
\newcommand{\lgla}{\longleftarrow}
\newcommand{\lgra}{\longrightarrow}
\newcommand{\lglra}{\longleftrightarrow}
\newcommand{\lan}{\langle}
\newcommand{\ran}{\rangle}
\newcommand{\llan}{\left\langle}
\newcommand{\rran}{\right\rangle}
\newcommand{\lce}{\left\lceil}
\newcommand{\rce}{\right\rceil}
\newcommand{\lfl}{\left\lfloor}
\newcommand{\rfl}{\right\rfloor}

\newcommand{\emphb}{\textcolor{blue}}
\newcommand{\emphg}{\textcolor{Grass}}
\newcommand{\emphr}{\textcolor{red}}
\newcommand{\eqFunc}{\overset{\mathrm{f}}{=}}
\newcommand{\eqDef}{\triangleq}
\newcommand{\diid}{\overset{\text{i.i.d.}}{\sim}}

\newcommand{\E}{\mathbb{E}}
\newcommand{\Var}{\mathsf{Var}}
\newcommand{\Cov}{\mathsf{Cov}}
\newcommand{\Bias}{\mathsf{Bias}}
\newcommand{\MSE}{\mathsf{MSE}}
\newcommand{\MLE}{\mathsf{MLE}}
\newcommand{\Risk}{\mathsf{R}}
\renewcommand{\Pr}{\mathbb{P}}
\newcommand{\Ber}{\mathrm{Ber}}
\newcommand{\Binom}{\mathrm{Binom}}
\newcommand{\Unif}{\mathrm{Unif}}

\maketitle

\begin{figure}[h]
    \centering
    \includegraphics[width=\columnwidth]{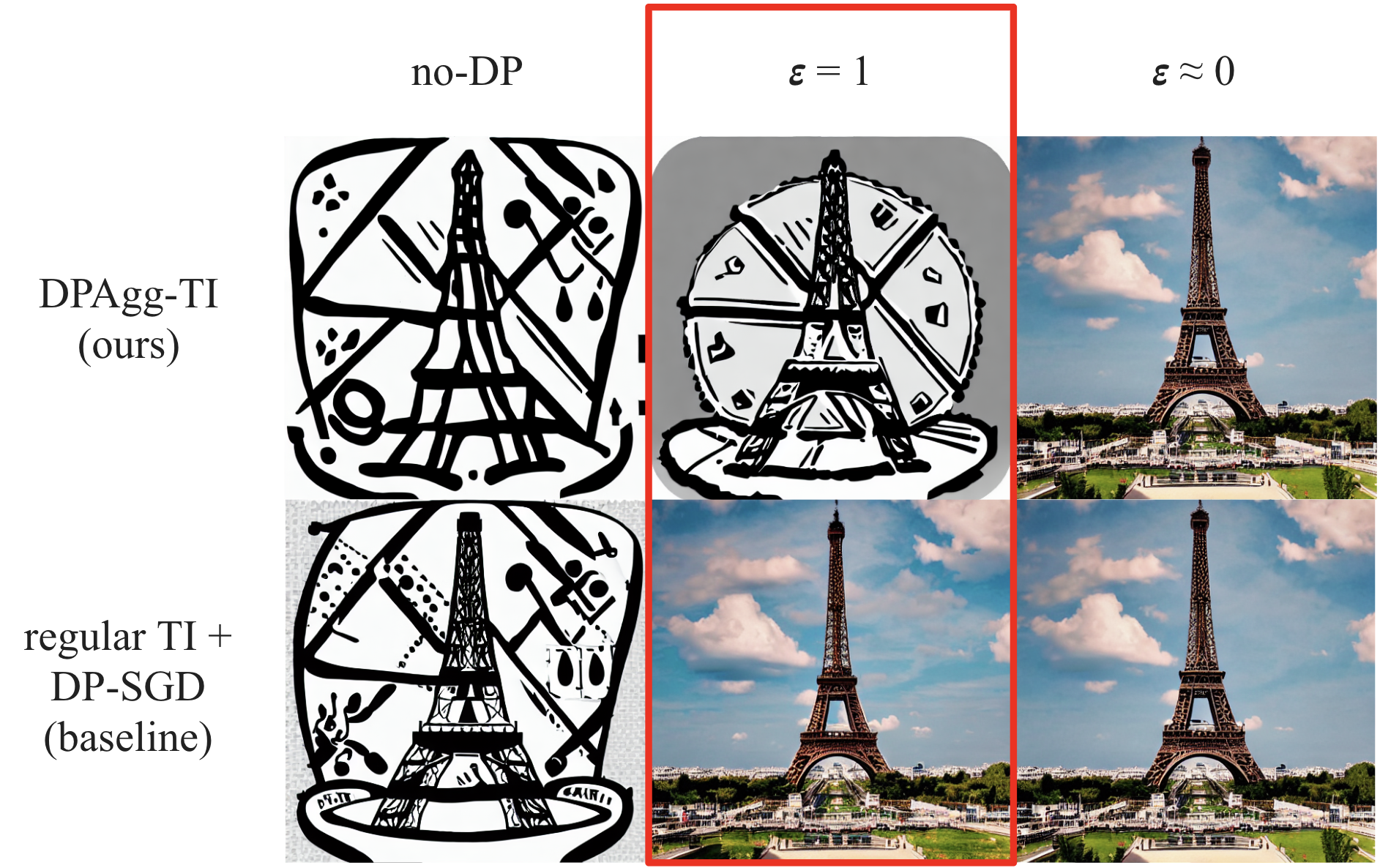}
    \caption{We compare our method (DPAgg-TI, top) to a baseline applying DP-SGD to Textual Inversion (bottom), using the prompt “an icon of the Eiffel Tower in the style of the Paris 2024 Olympic Pictograms.” While the baseline learns a single embedding over the dataset, our method privately aggregates per-image embeddings. At privacy budget $\varepsilon = 1$, DPAgg-TI preserves visual fidelity much better than the baseline, and closely matches the non-private output (left), demonstrating a superior privacy-utility tradeoff.
}
    \label{fig:one}
\end{figure}

\begin{figure*}[t] 
    \centering
    \includegraphics[width=\textwidth]{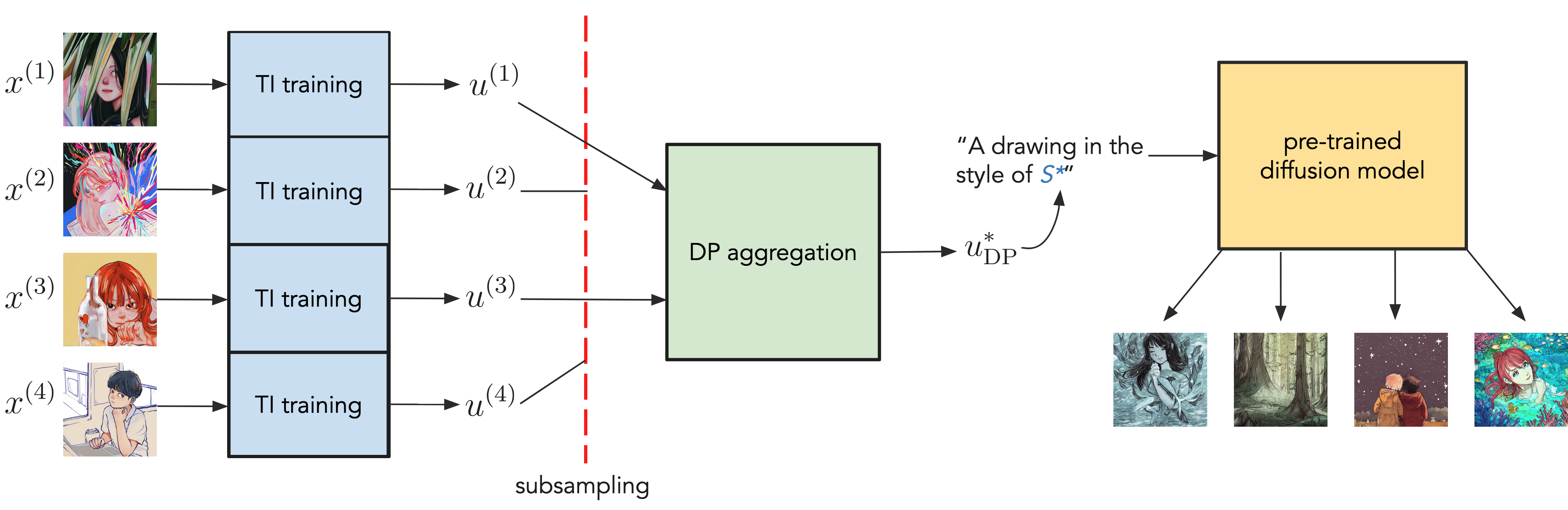}
    \caption{Overview of DPAgg-TI. We first apply Textual Inversion to extract embeddings for each image in the private dataset. 
    These embeddings are then aggregated with differentially private mechansim, incorporating subsampling to produce a private embedding $u^*_{\text{DP}}$.
    Finally, images are generated using the corresponding token \texttt{$S^*$}.}
    \label{fig:main}
\end{figure*}

\begin{abstract}
Personalizing large-scale diffusion models poses serious privacy risks, especially when adapting to small, sensitive datasets. A common approach is to fine-tune the model using differentially private stochastic gradient descent (DP-SGD), but this suffers from severe utility degradation due to the high noise needed for privacy, particularly in the small data regime. We propose an alternative that leverages Textual Inversion (TI), which learns an embedding vector for an image or set of images, to enable adaptation under differential privacy (DP) constraints. Our approach, Differentially Private Aggregation via Textual Inversion (DPAgg-TI), adds calibrated noise to the aggregation of per-image embeddings to ensure formal DP guarantees while preserving high output fidelity. We show that DPAgg-TI outperforms DP-SGD finetuning in both utility and robustness under the same privacy budget, achieving results closely matching the non-private baseline on style adaptation tasks using private artwork from a single artist and Paris 2024 Olympic pictograms. In contrast, DP-SGD fails to generate meaningful outputs in this setting.
\end{abstract}

\section{Introduction}
The rapid adoption of diffusion models ~\cite{ho2020denoising, song2020score, rombach2022high} has raised significant privacy and legal concerns.
These models are vulnerable to privacy attacks, such as membership inference~\cite{duan2023diffusion},
where attackers determine if a specific data point was used for training, and data extraction~\cite{carlini2023extracting},
which enables reconstruction of training data.
This risk is amplified during fine-tuning on smaller, domain-specific datasets, where each record has a greater impact.
Additionally, reliance on large datasets scraped without consent raises copyright concerns~\cite{vyas2023provable},
as diffusion models can reproduce original artworks without credit or compensation.
These issues highlight the urgent need for privacy-preserving technologies 
and clearer ethical and legal guidelines for generative models.

Differential privacy~(DP)~\cite{dwork2006differential} is a widely adopted framework for addressing these challenges. One standard approach for ensuring DP in deep learning is Differentially Private Stochastic Gradient Descent (DP-SGD)~\cite{abadi2016deep}, which modifies traditional SGD by adding noise to clipped gradients. However, applying DP-SGD to train diffusion models poses several challenges. It introduces significant computational and memory overhead due to per-sample gradient clipping \cite{hoory2021learning},
which is essential for bounding gradient sensitivity~\cite{dwork2006calibrating, abadi2016deep}.
DP-SGD is also incompatible with batch-wise operations like batch normalization,
as these link samples and hinder sensitivity analysis.
Furthermore, training large models with DP-SGD often leads to substantial performance degradation,
particularly under realistic privacy budgets since the required noise scales with the gradient norm. Consequently, existing diffusion models trained with DP-SGD are limited 
to small-scale images~\cite{dockhorn2023differentially, ghalebikesabi2023differentially}.

Independent of privacy concerns, Textual Inversion (TI)~\cite{gal2023an} effectively adapts diffusion models to specific styles or content without modifying the model. Instead, TI learns an external embedding vector that captures the style or content of a target image set, which is then incorporated into text prompts to guide the model’s outputs. A key advantage of TI is its ability to compress a style into a compact vector, reducing computational and memory demands while simplifying privacy mechanisms, as privacy constraints can be applied directly to embeddings rather than the full model. Additionally, since TI avoids direct model optimization, it remains efficient and compatible with DP constraints on smaller datasets.

In this work, we propose a novel privacy-preserving adaptation method for smaller datasets, leveraging TI to avoid the extensive model updates required by DP-SGD. Standard TI does not offer formal privacy guarantees, so we introduce a private variant, Differentially Private Aggregation via Textual Inversion (DPAgg-TI), summarized in Figure~\ref{fig:main}. Our method decouples interactions among samples by learning a separate embedding for each target image, which are then aggregated into a noisy centroid. This approach ensures efficient and secure adaptation to private datasets.

Our experiments demonstrate the effectiveness of DPAgg-TI,
showing that TI remains robust in preserving stylistic fidelity even under privacy constraints (Figure~\ref{fig:one}).
Applying our method to a private artwork collection by {\eve} and Paris 2024 Olympics pictograms~\cite{Paris2024Pictograms},
we show that DPAgg-TI captures nuanced stylistic elements while ensuring privacy.
We observe a trade-off between privacy (controlled by DP parameter $\varepsilon$) and image quality:
lower $\varepsilon$ reduces fidelity but maintains the target style under moderate noise.
Subsampling further amplifies privacy by reducing sensitivity to individual data points,
mitigating noise impact on image quality.
This framework enables privacy-preserving adaptation of diffusion models to new styles and domains 
while protecting sensitive data.

Our contributions can be summarized as follows:
\begin{itemize}
\item  We propose DPAgg-TI that ensures privacy by learning separate embeddings for individual images and aggregating them into a noisy centroid.
\item  Our approach enables style adaptation without extensive model updates, reducing computational overhead while preserving privacy.
\item  We analyze the trade-off between privacy and image quality, showing that moderate noise maintains stylistic fidelity while protecting sensitive data.
\item  We validate our method on diverse datasets, demonstrating its effectiveness in capturing stylistic elements under privacy constraints.
\end{itemize}

\section{Background and Related Work}
\subsection{Diffusion Models}

Diffusion models~\cite{ho2020denoising, song2020score, song2020denoising, rombach2022high} leverage an iterative denoising process to generate high-quality images that align with a given conditioning input from random noise. In text-to-image generation, this conditioning input is based on a textual description (a prompt) that guides the model in shaping the image to reflect the content and style specified by the text. To convert the text prompt into a suitable conditioning format, it is first broken down into discrete tokens, each representing a word or sub-word unit. These tokens are then converted into a sequence of embedding vectors $v_i$ that encapsulate the meaning of each token within the model’s semantic space. Next, these embeddings pass through a transformer text encoder, such as CLIP \cite{radford2021learning}, outputting a single text-conditioning vector $y$ that serves as the conditioning input. This vector $y$ is then incorporated at each denoising step, guiding the model to align the output image with the specific details outlined in the prompt.

The image generation process, also known as the reverse diffusion process, comprises of $T$ discrete timesteps and starts with pure Gaussian noise $x_T$. At each decreasing timestep $t$, the denoising model, which often utilizes a U-Net structure with cross-attention layers, takes a noisy image $x_t$ and text conditioning $y$ as inputs and predicts the noise component $\epsilon_\theta(x_t, y, t)$, where $\theta$ denotes the denoising model's parameters. The predicted noise is then used to make a reverse diffusion step from $x_t$ to $x_{t-1}$, iteratively refining the noisy image closer to a coherent output $x_0$ conditioned on $y$.

The objective function for a text-conditioned diffusion model, given both the noisy image $x_t$ and the text conditioning $y$, is typically a mean squared error between the true noise $\epsilon$ and the predicted noise $\epsilon_\theta(x_t, y, t)$. The denoising model is therefore trained over the optimization problem
\begin{equation}
  \theta^* = \arg\min_{\theta} \mathbb{E}_{x,\epsilon\sim\mathcal{N}(0,I),t\sim[T]}[\|\epsilon-\epsilon_\theta(x_t, y, t)\|^2].
  \label{eq:ldm_loss}
\end{equation}

\subsection{Textual Inversion} Textual Inversion (TI) \cite{gal2023an} is an adaptation technique that enables personalization using a small dataset of typically 3-5 images. This approach essentially learns a new token that encapsulates the semantic meaning of the training images, allowing the model to associate specific visual features with a custom token.

To achieve this, TI trains a new token embedding, denoted as $u$, representing a placeholder token, denoted as $S$. During training, images are conditioned on phrases such as ``A photo of $S$'' or ``A painting in the style of $S$''. However, unlike the fixed embeddings of typical tokens $v_i$, $u$ is a learnable parameter. Let $y_u$ denote the text conditioning vector resulting from a prompt containing the token $S$. Through gradient descent, TI minimizes the diffusion model loss with respect to $u$, instead of the diffusion model parameters $\theta$, which we keep fixed. By doing so, we iteratively refine this embedding to capture the unique characteristics of the training images. The resulting optimal embedding $u^*$ is formalized as 
\begin{equation}
  u^* = \arg\min_u \mathbb{E}_{x,\epsilon\sim\mathcal{N}(0,I),t\sim[T]}[\|\epsilon-\epsilon_\theta(x_t, y_u, t)\|^2].
  \label{eq:ti_loss}
\end{equation}
Hence, $u^*$ represents an optimized placeholder token $S^*$, which can employed in prompts such as ``A photo of $S^*$ floating in space'' or ``A drawing of a capybara in the style of $S^*$'', enabling the generation of personalized images that reflect the learned visual characteristics.

\subsection{Differential Privacy} In this work, we adopt differential privacy (DP) \cite{dwork2006calibrating, dwork2006differential} as our privacy framework. Over the past decade, DP has become the gold standard for privacy protection in both research and industry. It measures the stability of a randomized algorithm with respect to changes in an input instance, thereby quantifying the extent to which an adversary can infer the existence of a specific input based on the algorithm's output.

\begin{definition}[(Approximate) Differential Privacy]\label{def:DP}
For $\varepsilon, \delta \geq 0$, a randomized mechanism $\mcal{M}: \mcal{X}^n \ra \mcal{Y}$ satisfies $(\varepsilon, \delta)$-DP if for all \emph{neighboring} datasets $\mcal{D}, \mcal{D}' \in \mcal{X}^{n}$ which differ in a single record (i.e., $\lV \mcal{D} - \mcal{D}' \rV_{\msf{H}} \leq 1$ where $\lV \cdot \rV_\msf{H}$ is the Hamming distance) and all measurable $\mcal{S}$ in the range of $\mcal{M}$, we have that 
$$ \Pr\lp \mcal{M}(\mcal{D}) \in \mcal{S} \rp \leq e^\varepsilon \Pr\lp \mcal{M}(\mcal{D}') \in \mcal{S} \rp+\delta. $$
When $\delta = 0$, we say $\mcal{M}$ satisfies $\varepsilon$-pure DP or ($\varepsilon$-DP).
\end{definition}

To achieve DP, the Gaussian mechanism is commonly applied~\cite{dwork2014algorithmic,balle2018improving}, adding Gaussian noise scaled by the $\ell_2$-sensitivity $\Delta$ and privacy parameters $(\varepsilon,\delta)$. We add zero-mean isotropic Gaussian noise with standard deviation
\begin{equation}
  \sigma \;=\; \frac{\Delta\,\sqrt{2\ln(1.25/\delta)}}{\varepsilon}.
  \label{eq:gaussian_mech_sigma}
\end{equation}
In practice, we calibrate $\sigma$ using numerical privacy accountants (e.g., the analytic Gaussian mechanism and RDP)~\cite{balle2018improving,mironov2017renyi}. This mechanism enables a smooth privacy–utility trade-off and is widely used in privacy-preserving machine learning, including DP-SGD~\cite{abadi2016deep}, which adds Gaussian noise to model updates to achieve DP.

\subsubsection{Privacy Amplification by Subsampling} 
\label{sec:subsampling} Subsampling is a standard technique in DP, where a full dataset of size \( n \) is first subsampled to \( m \) records without replacement (typically with \( m \ll n \)) before the privatization mechanism (e.g. the Gaussian mechanism) is applied. Specifically, if a mechanism provides \((\varepsilon, \delta)\)-DP on a dataset of size \( m \), it achieves \((\varepsilon', \delta')\)-DP on the subsampled dataset, where \(\delta' = \frac{m}{n} \delta\) and 
\begin{equation}
    \varepsilon' = \log \left( 1 + \frac{m}{n} \left( e^\varepsilon - 1 \right) \right) = O \left( \frac{m}{n} \varepsilon \right).
\label{eq:pabs}
\end{equation}
This result is well-known \cite[Theorem~29]{steinke2022composition},
with tighter amplification bounds available for Gaussian mechanisms~\cite{mironov2017renyi}.

\subsection{Private Adaptation of Diffusion Models}
Recent advancements in applying DP to diffusion models have aimed to balance privacy preservation 
with the high utility of generative outputs. Early work on differentially private generative models, such as Chen et al.'s~\cite{chen2021differentially} investigation of DP-GANs with model inversion defenses,
established foundational principles for protecting generative models from privacy breaches during training.
Dockhorn et al.~\cite{dockhorn2023differentially} proposed a Differentially Private Diffusion Model (DPDM)
that enables privacy-preserving generation of realistic samples,
setting a foundational approach for adapting diffusion processes using DP-SGD.
Another common strategy involves training a model on a large public dataset,
followed by differentially private fine-tuning on a private dataset \cite{ghalebikesabi2023differentially}.
While effective in certain contexts, this approach raises privacy concerns,
particularly around risks of information leakage during the fine-tuning phase~\cite{tramer2024position}.

In response to these limitations, various adaptation techniques have emerged.
Although not specific to diffusion models, some methods focus on training models on synthetic data followed 
by DP-constrained fine-tuning, as in Yu et al.~\cite{yu2024vip},
which demonstrates the feasibility of applying DP in later adaptation stages.
Other approaches explore differentially private learning of feature representations~\cite{sander2024differentially},
aiming to distill private information into a generalized embedding space while maintaining DP guarantees.
Although these adaptations are not yet implemented for diffusion models,
they lay essential groundwork for developing secure and efficient privacy-preserving generative models.

\begin{figure*}[!ht]
\centering
\includegraphics[width=0.97\textwidth]{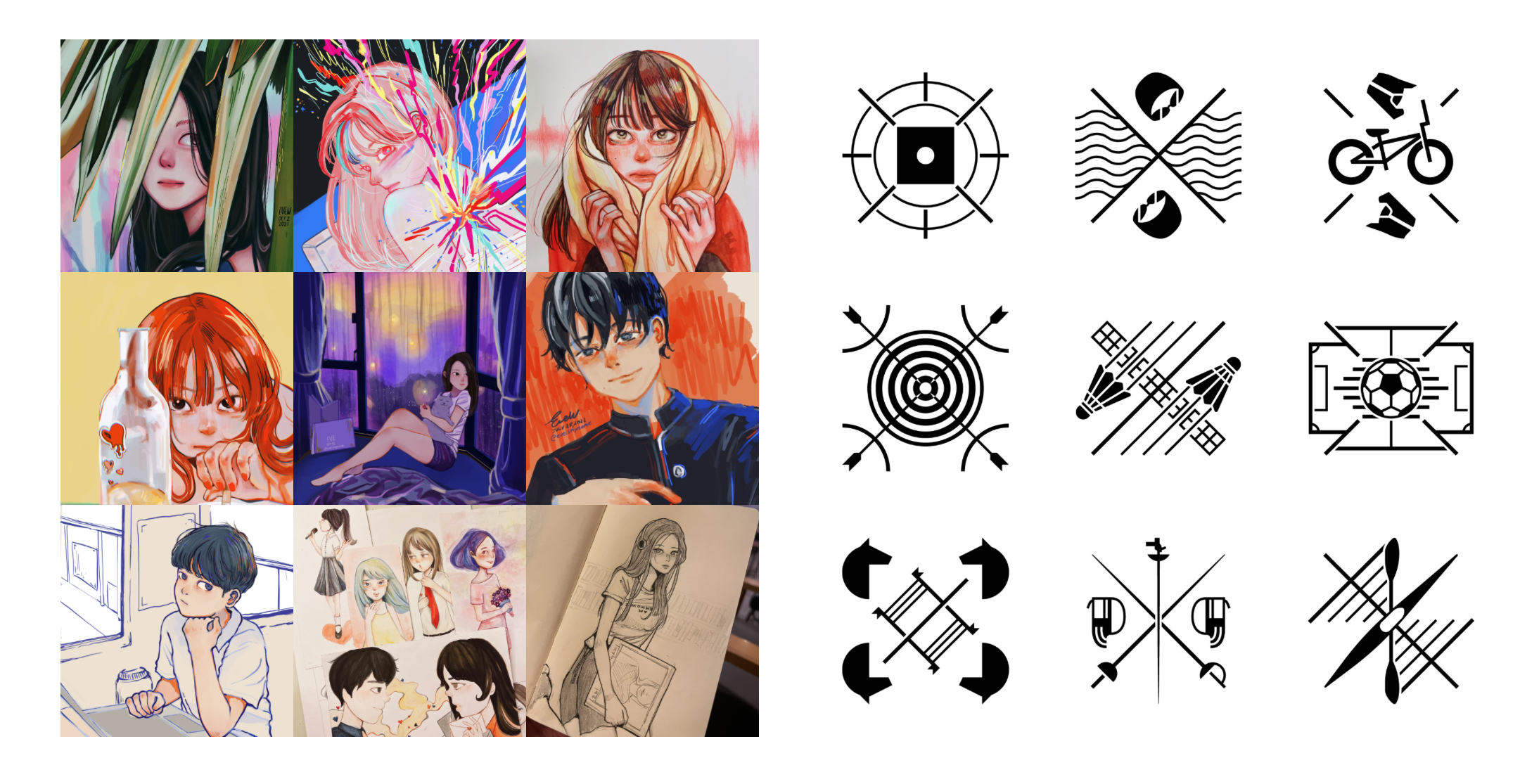}
\caption{Samples of images used in our style adaptation experiments. \textbf{Left:} artwork by {\eve} ($n = 158$). \textbf{Right:} Paris 2024 Olympic pictograms ($n = 47$), \textcopyright~\textit{International Olympic Committee, 2023}.}
\label{fig:datasets}
\end{figure*}

\section{Differentially Private Adaptation via Textual Inversion}

Let $x^{(1)}, \dots, x^{(n)}$ represent a target dataset of images whose characteristics we wish to privately adapt our image generation towards.
Instead of training a single token embedding on the entire dataset as in regular TI,
we train a separate embedding $u^{(i)}$ on each $x^{(i)}$ to obtain a set of embeddings $u^{(1)}, \dots, u^{(n)}$, as illustrated in Figure \ref{fig:main}.
We can formalize the encoding process as 
\begin{equation}
  u^{(i)} = \arg\min_u \mathbb{E}_{\epsilon\sim\mathcal{N}(0,I),t}[\|\epsilon-\epsilon_\theta(x_t^{(i)}, y_u, t)\|^2].
\label{eq:indiv_loss}
\end{equation}
\pagebreak

Then, we aggregate the embeddings $u^{(1)}, \dots, u^{(n)}$ by calculating their centroid and adding isotropic Gaussian noise. 
To ensure bounded sensitivity, we employ a purely directional token embedding (semantics depend only on direction), such as CLIP~\cite{radford2021learning}, and $\ell_2$-normalize each embedding vector prior to aggregation. 
We can therefore define the resulting centroid $u^*_{\text{DP}}$ as
\begin{equation}
  u^*_{\text{DP}} = \frac{1}{n}\sum_{i=1}^{n}\frac{u^{(i)}}{\|u^{(i)}\|} + \mathcal{N}(0, \sigma^2I),
  \label{eq:centroid_method}
\end{equation}
where $\sigma$ is given by~\eqref{eq:gaussian_mech_sigma}. 

Under this normalization, the $\ell_2$-sensitivity of $u^*_{\text{DP}}$ is
\begin{equation}
\Delta = \frac{1}{n}\sup_{u,u'}\left\|\frac{u}{\|u\|} - \frac{u'}{\|u'\|}\right\| = \frac{2}{n},
\label{eq:sensitivity}
\end{equation}
The noisy centroid embedding $u^*_{\text{DP}}$ can then be used to adapt the downstream image generation process. 
Similar to regular TI's $u^*$, we can use $u^*_{\text{DP}}$ to represent a new placeholder token $S^*$ 
that can be incorporated into prompts for personalized image generation.
While $u^*_{\text{DP}}$ may not fully solve the TI optimization problem presented in \eqref{eq:ti_loss},
it provides provable privacy guarantees, with only a minimal trade-off in accurately 
representing the style of the target dataset.

To reduce the amount of noise needed to provide the same level of DP, we employ subsampling:
instead of computing the centroid over all $n$ embedding vectors,
we randomly sample $m \le n$ embedding vectors without replacement and compute the centroid over only the sampled vectors.
Then the standard privacy amplification by subsampling bounds (such as \eqref{eq:pabs}) can be applied.
Formally, we sample $\Dsub \subseteq \{u^{(1)}, \dots, u^{(n)}\}$ where $|\Dsub| = m$,
and compute $u^*_{\text{DP}}$ as 
\begin{equation}
  u^*_{\text{DP}} = \frac{1}{m}\sum_{u^{(i)} \in \Dsub}\frac{u^{(i)}}{||u^{(i)}||} + \mathcal{N}(0, \sigma^2I),
  \label{eq:centroid_subsample}
\end{equation}
where $\sigma$ can be computed numerically using~\eqref{eq:gaussian_mech_sigma} with $\varepsilon'$ and $\delta'$ from Section \ref{sec:subsampling}.

\section{Experimental Results}
\subsection{Datasets}
We compiled two datasets to evaluate our style adaptation method, specifically selecting content unlikely to be recognized by Stable Diffusion v1.5, our base model.

The first dataset consists of 158 artworks by the artist {\eve}, who has granted consent for non-commercial use.
This dataset allows us to assess whether models can capture artistic styles without memorizing individual works.
While some of these artworks may have been publicly accessible on social media, making incidental inclusion in Stable Diffusion’s pretraining possible,
the artist’s limited recognition and relatively small portfolio reduce the likelihood that the model has internalized her unique style.
This dataset serves as a controlled test for privacy-preserving style transfer on individual artistic collections.

The second dataset contains 47 pictograms from the Paris 2024 Olympics~\cite{Paris2024Pictograms}, permitted strictly for non-commercial editorial use~\cite{IOCProperties}.
These pictograms were officially released in February 2023, several months after the release of Stable Diffusion v1.5, ensuring they were absent from the model’s pretraining data.
This dataset allows us to assess how well our approach adapts to newly introduced visual styles that the base model has never encountered.

Both datasets are used to test the ability of our method to extract and transfer stylistic elements while preserving privacy.
Representative samples are shown in Figure~\ref{fig:datasets}.

\begin{figure*}[h]
\centering
\includegraphics[width=0.95\textwidth]{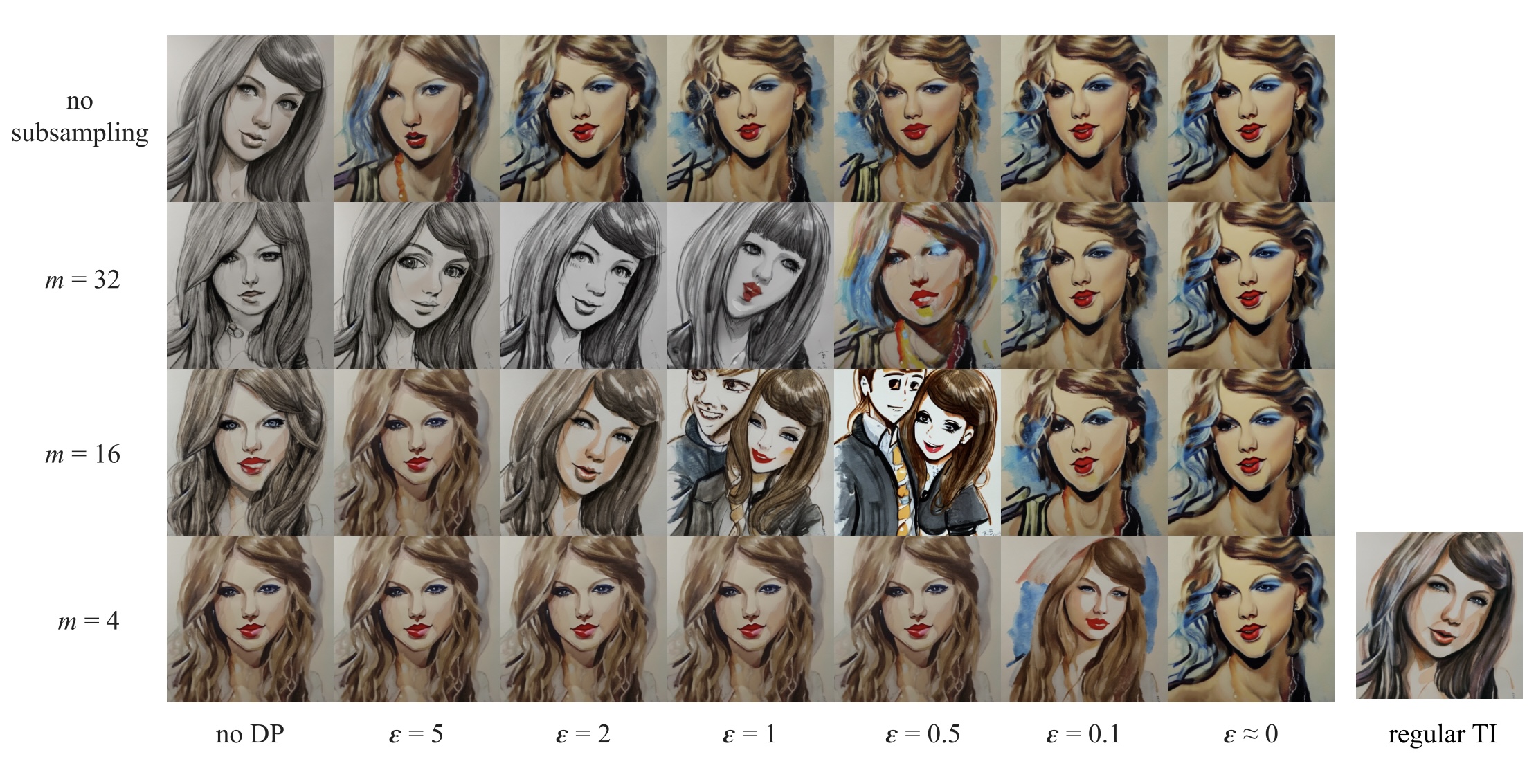}
\caption{Images generated by Stable Diffusion v1.5 using the prompt ``A painting of Taylor Swift in the style of $<${\eve}$>$'', 
with the embedding $<${\eve}$>$ trained using different values of $m$ and $\varepsilon$.}
\label{fig:eveismyname}
\end{figure*}
\begin{figure*}[h]
\centering
\includegraphics[width=0.95\textwidth]{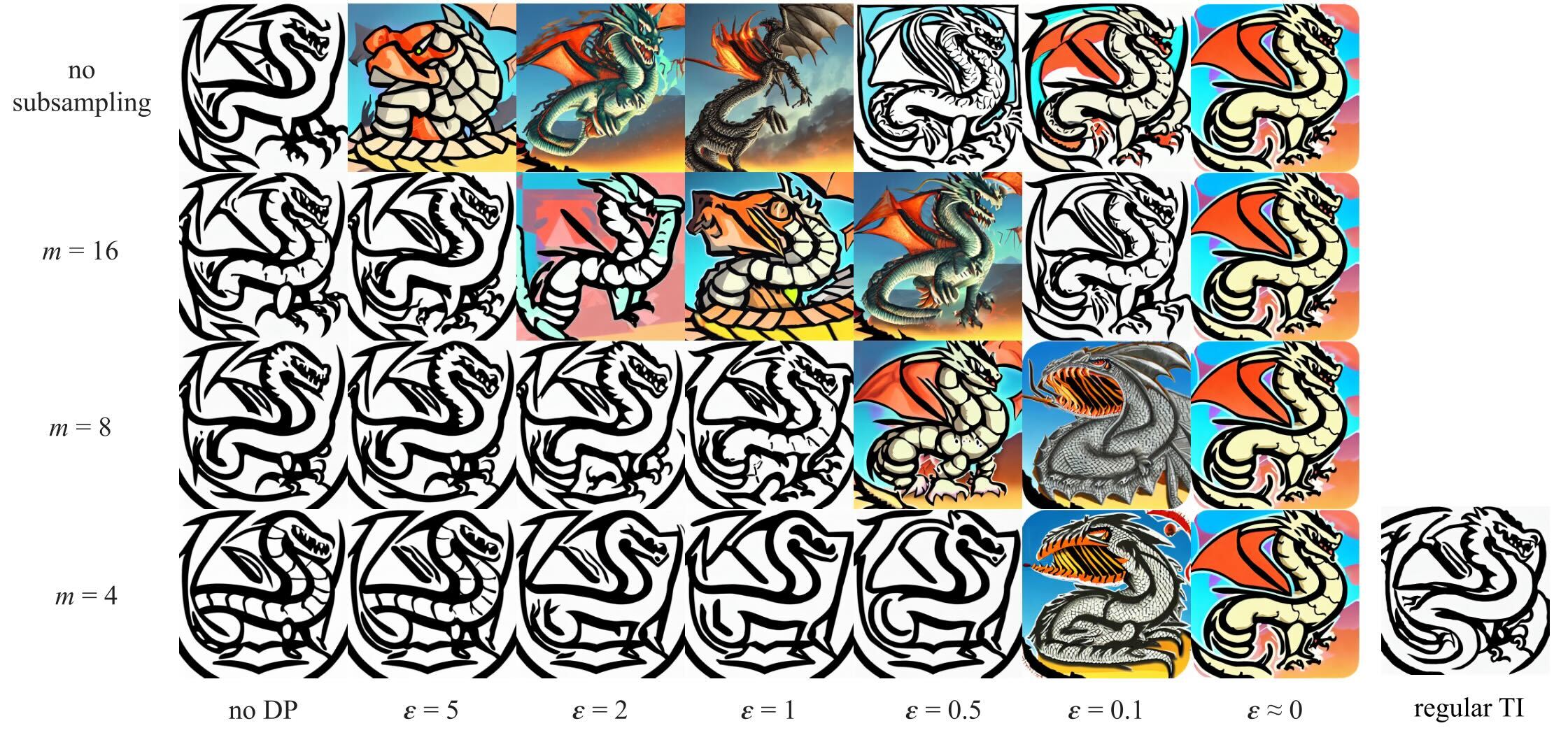}
\caption{Images generated by Stable Diffusion v1.5 using the prompt ``Icon of a dragon in the style of $<$Paris 2024 Pictograms$>$'',
with the embedding $<$Paris 2024 Pictograms$>$ trained using different values of $m$ and $\varepsilon$.}
\label{fig:paris}
\end{figure*}

\subsection{Style Transfer Results}

Using both the {\eve} and Paris 2024 pictograms dataset, we trained TI~\cite{gal2023an} embeddings on Stable Diffusion v1.5~\cite{rombach2022high} using DPAgg-TI. Our primary goal is to investigate how DP configurations, specifically the privacy budget $\varepsilon$ and subsampling size $m$, affect the generated images quality and privacy resilience. For regular TI, we utilize the default process to embed the private dataset without any additional noise.
For the DPAgg-TI, we test multiple configurations of $m$ and $\varepsilon$ to analyze the trade-off between image fidelity and privacy.%

Figures~\ref{fig:eveismyname} and~\ref{fig:paris} present generated images across two key configurations:
(1) regular TI without DP, (2) DPAgg-TI with DP at different values of $m$ and $\varepsilon$.
To ensure reproducibility and fair comparison across all experimental conditions, we fixed the random seed for the entire generation pipeline. This design choice allows us to isolate the effect of our style transfer method while holding other sources of randomness constant.  As with common practice, we set $\delta = 1/n$.

Images generated without DP closely resemble the unique stylistic elements of the target dataset. In particular, images adapted using {\eve} images displayed crisp details and nuanced color gradients characteristic of the artist's work (Figure~\ref{fig:eveismyname}), while those of Paris 2024 pictograms captured the logo’s original structure (Figure~\ref{fig:paris}).
In contrast, DP configurations introduce a discernible degradation in image quality, with lower epsilon values and smaller subsampling sizes resulting in diminished stylistic fidelity.

As a no-learning baseline, we consider the limit $\varepsilon \to 0$, under which $u^*_{\text{DP}}$ should convey zero information about the target dataset. Since $\sigma \propto 1/\varepsilon$ is undefined at $\varepsilon=0$, we approximate this regime by setting $\varepsilon=10^{-5} \approx 0$, which yields effectively infinite noise.

As $\varepsilon \rightarrow 0$, the resulting token embedding $u^*_{\text{DP}}$ gradually loses its semantic meaning, leading to a loss of stylistic fidelity. In particular, $y_{u^*_{\text{DP}}}$ tends towards $y$ (a conditioning vector independent of the learnable embedding). In our results, this manifests as a painting of Taylor Swift devoid of the artist-specific stylistic elements (Figure~\ref{fig:eveismyname}), or a generic icon of a dragon (with color, as opposed to the black and white design of the pictograms, Figure~\ref{fig:paris}). To verify this interpretation, we generated images with the same prompts but without the special token $S^*$ and compared them to the $\varepsilon \approx 0$ generations. The images were visually identical, confirming that at $\varepsilon \approx 0$, the token becomes semantically meaningless and is ignored by the text encoder.

With this in mind, given a fixed seed, $\varepsilon$ can be interpreted as a drift parameter, representing the progression from the optimal $u^*_{\text{DP}}$ towards a semantically meaningless embedding, gradually steering the generated image away from the target style in exchange for stronger privacy guarantees. We also observe instances where there is a temporary drop in prompt fidelity (e.g., $m = 16, \varepsilon \in [0.5, 1]$ in Figure \ref{fig:eveismyname} and intermediate $\varepsilon$ values in Figure \ref{fig:paris}) which restores as $u^*_{\text{DP}}$ drifts even further from its optimal value. We hypothesize that this is due to drifted $u^*_{\text{DP}}$ capturing a different meaning unrelated to the prompt, before losing any meaning that could be interpreted by Stable Diffusion's text encoder, causing $u^*_{\text{DP}}$ to be disregarded from $y_{u^*_{\text{DP}}}$ and the prompt fidelity to be restored. 

Meanwhile, reducing $m$ also reduces the sensitivity of the generated image to with respect to $\varepsilon$, as evident by the observation that, on both datasets at $m = 4$, (subsampling rate below $0.1$) image generation can tolerate $\varepsilon$ as low as $0.5$ without significant changes in visual characteristics, and retaining stylistic elements of the target dataset at $\varepsilon$ as low as $0.1$. This strong boost in robustness comes at a small price of base style capture fidelity. As observed in Figures \ref{fig:eveismyname} and \ref{fig:paris}, we can also treat subsampling as an introduction of noise. Mathematically, the subsample centroid is an unbiased estimate of the true centroid, and so the subsampling process itself defines a distribution centered at the true centroid. However, the amount of noise introduced by the subsampling process is limited by the individual image embeddings, as a subsample centroid can only stray from the true centroid as much as the biggest outlier in the dataset. 
\begin{figure*}[!h]
\centering
\includegraphics[width=\textwidth]{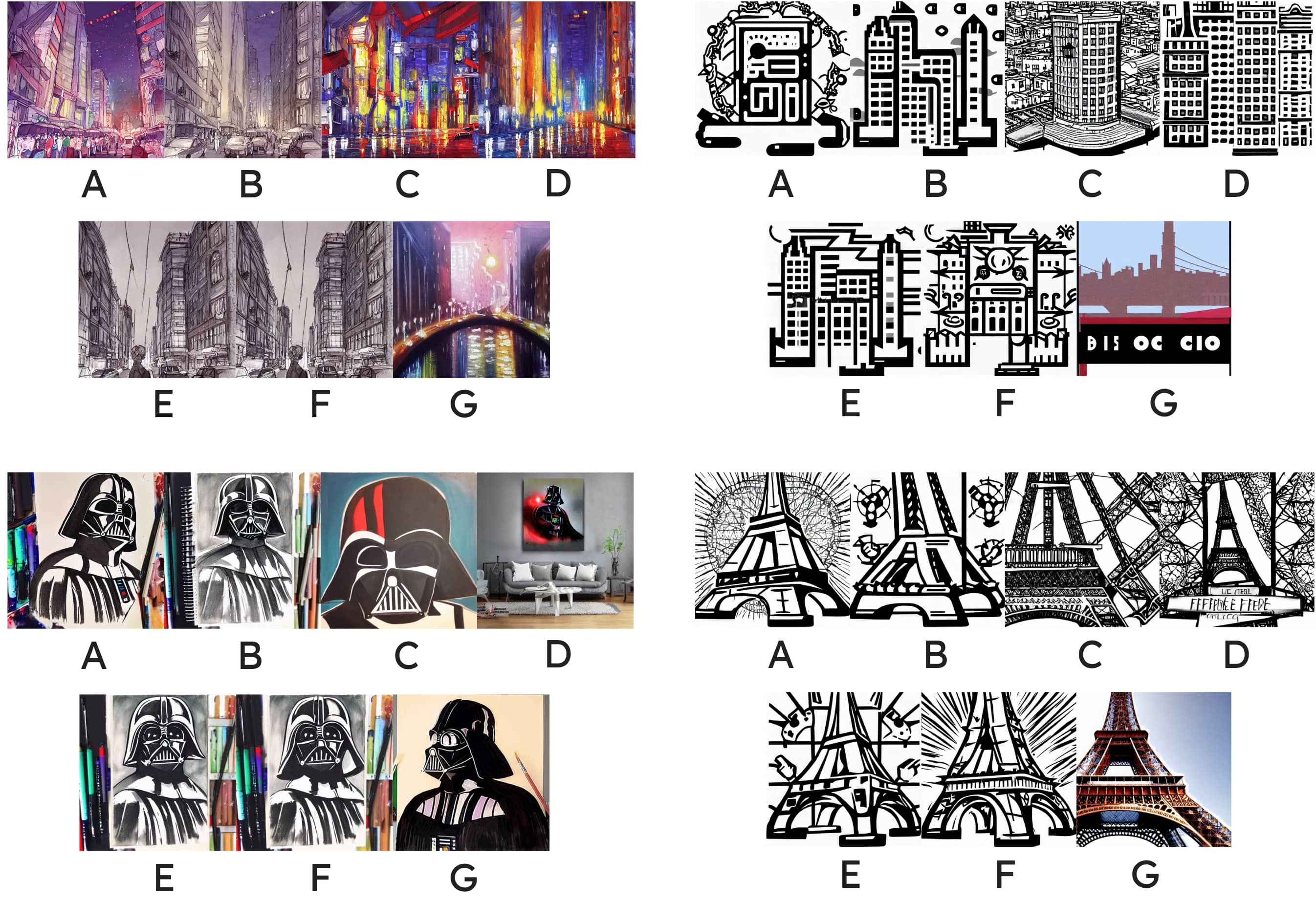}
\caption{Samples of image sets used in our user study. Participants were asked to compare 2 images at a time.}
\label{fig:questionaire}
\end{figure*}

\begin{table*}[h!]
\centering 
\caption{Survey Results.}
\label{tab:survey_res}
\begin{tabular}{|p{2.7cm}|p{5.2cm}|p{5.2cm}|p{1.7cm}|} 
\hline
         & Regular TI  &  No Adaptation  & Unsure \\ \hline
\textit{\eve}   & 19 & 4  & 2      \\ \hline
\textit{Paris 2024} & 16 & 6  & 3      \\ \hline
\end{tabular}
\end{table*}

\begin{table*}[h!]
\centering
\begin{tabular}{|p{2.7cm}|p{5.2cm}|p{5.2cm}|p{1.7cm}|} 
\hline
         & DPAgg-TI (no DP, no subsampling)  & No Adaptation  & Unsure \\ \hline
\textit{\eve}   & 16 & 9  & 0      \\ \hline
\textit{Paris 2024} & 15 & 4 & 6      \\ \hline
\end{tabular}
\end{table*}

\begin{table*}[h!]
\centering
\begin{tabular}{|p{2.7cm}|p{5.2cm}|p{5.2cm}|p{1.7cm}|} 
\hline
         & Regular TI  & DPAgg-TI (no DP, no subsamp.)   & Unsure \\ \hline
\textit{\eve}   & 12 & 13  & 0      \\ \hline
\textit{Paris 2024} & 9 & 10 & 6      \\ \hline
\end{tabular}
\end{table*}

\begin{table*}[h!]
\centering
\begin{tabular}{|p{2.7cm}|p{5.2cm}|p{5.2cm}|p{1.7cm}|} 
\hline
         & Regular TI  & DPAgg-TI (no DP, subsamp.\ $m = 8$)  & Unsure \\ \hline
\textit{\eve}   & 16 & 6  & 3      \\ \hline
\textit{Paris 2024} & 7 & 13 & 5      \\ \hline
\end{tabular}
\end{table*}

\begin{table*}[h!]
\centering
\begin{tabular}{|p{2.7cm}|p{5.2cm}|p{5.2cm}|p{1.7cm}|} 
\hline
         & DPAgg-TI (no DP, no subsampling)  &  DPAgg-TI (no DP, subsamp.\ $m = 8$)  & Unsure \\ \hline
\textit{\eve}   & 18 & 4  & 3      \\ \hline
\textit{Paris 2024} & 10 & 8 & 7      \\ \hline
\end{tabular}
\end{table*}

\begin{table*}[h!]
\centering
\begin{tabular}{|p{2.7cm}|p{5.2cm}|p{5.2cm}|p{1.7cm}|} 
\hline
         & DPAgg-TI ($\varepsilon = 1$) no subsampling  & DPAgg-TI ($\varepsilon = 1$, subsamp.\ $m = 8$)  & Unsure \\ \hline
\textit{\eve}   & 14 & 10  & 1      \\ \hline
\textit{Paris 2024} & 3 & 16  & 6      \\ \hline
\end{tabular}
\end{table*}

\begin{table*}[h!]
\centering
\begin{tabular}{|p{2.7cm}|p{5.2cm}|p{5.2cm}|p{1.7cm}|} 
\hline
         &  DPAgg-TI (no DP, no subsampling)  & Style Guidance  & Unsure \\ \hline
\textit{\eve}   & 16 & 8  & 1      \\ \hline
\textit{Paris 2024} & 20 & 2  & 3      \\ \hline
\end{tabular}
\end{table*}

\begin{table*}[h!]
\centering
\begin{tabular}{|p{2.7cm}|p{5.2cm}|p{5.2cm}|p{1.7cm}|} 
\hline
         & DPAgg-TI ($\varepsilon = 1$, subsamp.\ $m = 8$) & Style Guidance  & Unsure \\ \hline
\textit{\eve}   & 16 & 8  & 1      \\ \hline
\textit{Paris 2024} & 19 & 2  & 4      \\ \hline
\end{tabular}
\end{table*}

\begin{table*}[h!]
\centering
\begin{tabular}{|p{2.7cm}|p{5.2cm}|p{5.2cm}|p{1.7cm}|} 
\hline
         &     DPAgg-TI (no DP, subsamp.\ $m = 8$) & DPAgg-TI ($\varepsilon = 1$, subsamp.\ $m = 8$)  & Unsure \\ \hline
\textit{\eve}   & 8 & 5  & 12      \\ \hline
\textit{Paris 2024} & 15 & 4  & 6      \\ \hline
\end{tabular} 
\end{table*}

\subsection{User Study}
\label{sec:userstudy}

To evaluate our approach under different DP and subsampling configurations, 
we conducted a user study with 25 participants. 
The goal was to assess whether DPAgg-TI preserves perceptual quality while offering privacy guarantees.

\subsubsection{Study Design and Setup}
Participants were shown reference images from two datasets: the {\eve} dataset of private artwork and the Paris 2024 Pictogram dataset. 
For each dataset, we used 10 prompts to generate images, resulting in 20 groups in total. 
Each group included images produced under different configurations, including regular TI, DPAgg-TI with and without DP noise, with and without subsampling ($m=8$), style guidance (see Appendix \ref{app:Style Guidance}), and a no-adaptation baseline.

\subsubsection{Survey Procedure}
Each participant evaluated two groups, one randomly selected from each dataset. 
For each group, participants were first shown the reference images, then asked to compare pairs of generated images produced with different configurations (see Figure~\ref{fig:questionaire}).
For each pair, they indicated which image better captured the reference style, or marked the choice as ``unsure.''

\subsubsection{Results and Analysis}
Survey results are summarized in Table~\ref{tab:survey_res}. 
Participants showed no clear preference between regular TI and DPAgg-TI, suggesting that our privacy-preserving approach maintains perceptual quality. 
As expected, both DP noise and smaller subsampling size degraded style fidelity, consistent with the trade-offs inherent in differential privacy. 
At $\varepsilon=1$, preferences were split between configurations with and without subsampling, although the subsampling variant was generally favored.
\pagebreak

Overall, the findings highlight that DPAgg-TI achieves perceptual quality comparable to regular TI, while subsampling serves as an effective mechanism to balance privacy and stylistic fidelity.

\subsection{Kernel Inception Distance}

The Kernel Inception Distance (KID)~\cite{binkowski2018demystifying} is a metric for evaluating generative models by measuring 
the difference between the distributions of generated and training images in an embedding space.
To compute KID, images generated by the model and real training images are passed through an Inception network~\cite{szegedy2016rethinking},
and their distributional differences are estimated.
Unlike the more commonly used Fréchet Inception Distance (FID)~\cite{heusel2017gans},
KID is an unbiased estimator of the true divergence between the learned and target distributions~\cite{Jayasumana_2024}, making it more suitable for smaller datasets, as in our case.
\begin{table*}[h!]
\caption{KID scores of DPAgg-TI on {\eve} dataset for various $\varepsilon$ values ranging 
from $\varepsilon =10^{-5}, 0.1, 0.5, 1.0, 5.0$ (including no DP)
under different subsampling levels ($m=4, 8, 16, 32$) as well as regular TI (\texttt{ctrl}). Reported values are the mean $\pm$ standard deviation over 100 random subsamples.}
\centering
\resizebox{\textwidth}{!}{
\begin{tabular}{l|rrrrrr}
\toprule
   $m$ &  No DP & $\varepsilon=$ 5.0 & $\varepsilon=$ 1.0 & $\varepsilon=$ 0.5 & $\varepsilon=$ 0.1 & $\varepsilon\approx 0$ \\
\midrule
-- & 0.0441 ± 0.0027 & 0.0798 ± 0.0032 & 0.0526 ± 0.0022 & 0.0688 ± 0.0020 & 0.1114 ± 0.0032 & 0.0654 ± 0.0027 \\
32 & 0.0753 ± 0.0047 & 0.0836 ± 0.0042 & 0.1166 ± 0.0037 & 0.0295 ± 0.0019 & 0.0644 ± 0.0021 & 0.0650 ± 0.0025 \\
16 & 0.0350 ± 0.0020 & 0.0381 ± 0.0018 & 0.0663 ± 0.0025 & 0.1303 ± 0.0033 & 0.0438 ± 0.0030 & 0.0660 ± 0.0029 \\
8  & 0.0359 ± 0.0018 & 0.0364 ± 0.0017 & 0.0366 ± 0.0019 & 0.0394 ± 0.0025 & 0.0527 ± 0.0033 & 0.0654 ± 0.0024 \\
4  & 0.0246 ± 0.0013 & 0.0251 ± 0.0016 & 0.0249 ± 0.0014 & 0.0256 ± 0.0012 & 0.0313 ± 0.0017 & 0.0653 ± 0.0023 \\
\texttt{ctrl} & 0.0314 ± 0.0010 & -- & -- & -- & -- & -- \\
\bottomrule
\end{tabular}}
\label{kids1}
\end{table*}
\begin{table*}[h!]
\centering
\caption{KID scores of DPAgg-TI on Paris dataset for various $\varepsilon$ values ranging 
from $\varepsilon =10^{-5}, 0.1, 0.5, 1.0, 5.0$ (including no DP)
under different subsampling levels ($m=4, 8, 16, 32$) as well as regular TI (\texttt{ctrl}). Reported values are the mean $\pm$ standard deviation over 100 random subsamples.}
\resizebox{\textwidth}{!}{
\begin{tabular}{l|rrrrrr}
\toprule
   $m$ &  No DP & $\varepsilon=$ 5.0 & $\varepsilon=$ 1.0 & $\varepsilon=$ 0.5 & $\varepsilon=$ 0.1 & $\varepsilon\approx 0$ \\
\midrule
-- & 0.1153 ± 0.0055 & 0.1194 ± 0.0054 & 0.1306 ± 0.0046 & 0.1395 ± 0.0057 & 0.1201 ± 0.0053 & 0.1274 ± 0.0055 \\
32 & 0.1222 ± 0.0066 & 0.1036 ± 0.0065 & 0.1375 ± 0.0047 & 0.1311 ± 0.0048 & 0.1248 ± 0.0060 & 0.1258 ± 0.0054 \\
16 & 0.1321 ± 0.0057 & 0.1411 ± 0.0077 & 0.1309 ± 0.0061 & 0.1380 ± 0.0047 & 0.1359 ± 0.0060 & 0.1273 ± 0.0057 \\
8  & 0.1303 ± 0.0084 & 0.1303 ± 0.0074 & 0.1112 ± 0.0062 & 0.1311 ± 0.0064 & 0.1318 ± 0.0052 & 0.1267 ± 0.0056 \\
4  & 0.1158 ± 0.0057 & 0.1085 ± 0.0056 & 0.1184 ± 0.0068 & 0.1194 ± 0.0065 & 0.1592 ± 0.0065 & 0.1268 ± 0.0055 \\
\texttt{ctrl} & 0.1383 ± 0.0066 & -- & -- & -- & -- & -- \\
\bottomrule
\end{tabular}}
\label{kids2}
\end{table*}

We report KID scores for different parameters in Tables ~\ref{kids1} and \ref{kids2}. Our results indicate that DPAgg-TI preserves the style transfer fidelity of TI while also ensuring differential privacy. Notably, for {\eve} ($m = 4$) at low privacy budgets, we observe even lower KID values than standard TI, suggesting enhanced style alignment. Similarly, results for the Paris 2024 dataset follow a comparable trend, with DPAgg-TI achieving KID scores similar to TI at low privacy budgets. However, the overall KID scores for this dataset remain high within the context of diffusion model style transfer.

\begin{figure}[h]
\centering
\includegraphics[width=\columnwidth]{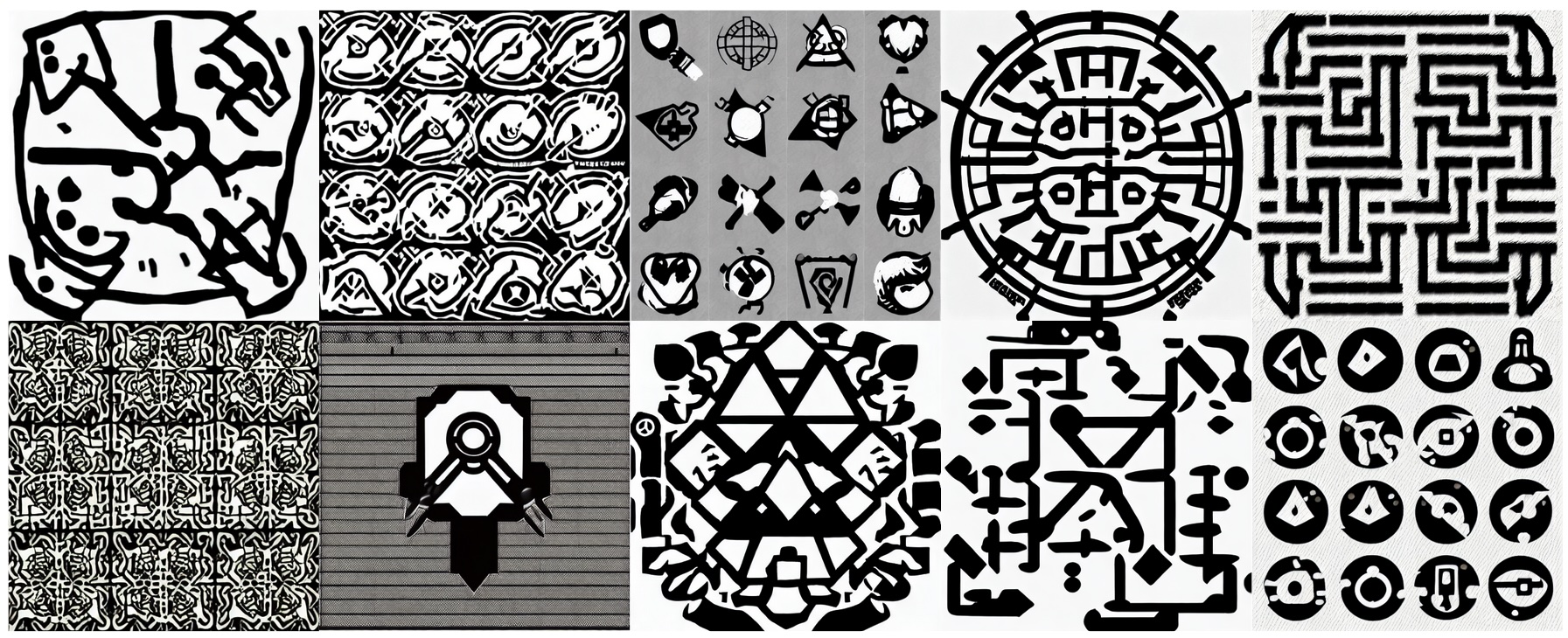}
\caption{Sample of generated images for KID evaluations with respect to the Paris 2024 dataset.}
\label{fig:paris-kids}
\end{figure}

Upon inspecting the generated images (Figure~\ref{fig:paris-kids}), we hypothesize that the abstract and out-of-distribution nature of the Paris 2024 images poses a challenge for the Inception network, leading to less meaningful feature embeddings. This likely inflates the measured embedding distances between generated and reference images, resulting in unusually high KID values.

For KID evaluations, we used prompts similar to those employed during TI training: ``A painting/icon in the style of $S^*$''. Consistent with the training image captions, these prompts do not specify a subject. For each parameter configuration, we generate 100 images and compute KID by repeatedly subsampling the larger of the real and generated sets to match the size of the smaller set 100 times, then averaging the resulting KID scores.

\pagebreak

\begin{figure}[h]
    \centering
    \includegraphics[width=\columnwidth]{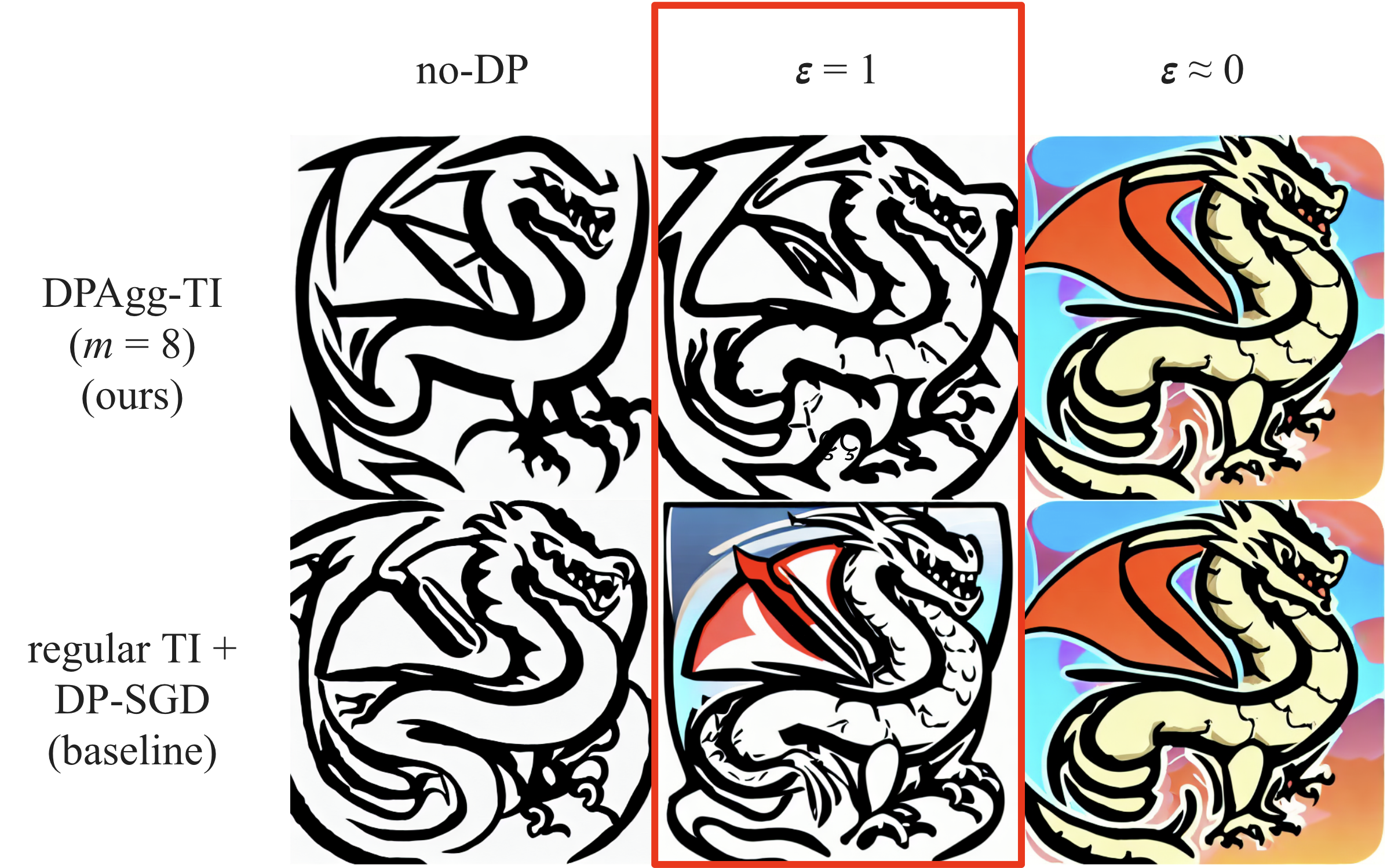}
    \includegraphics[width=0.99\columnwidth]{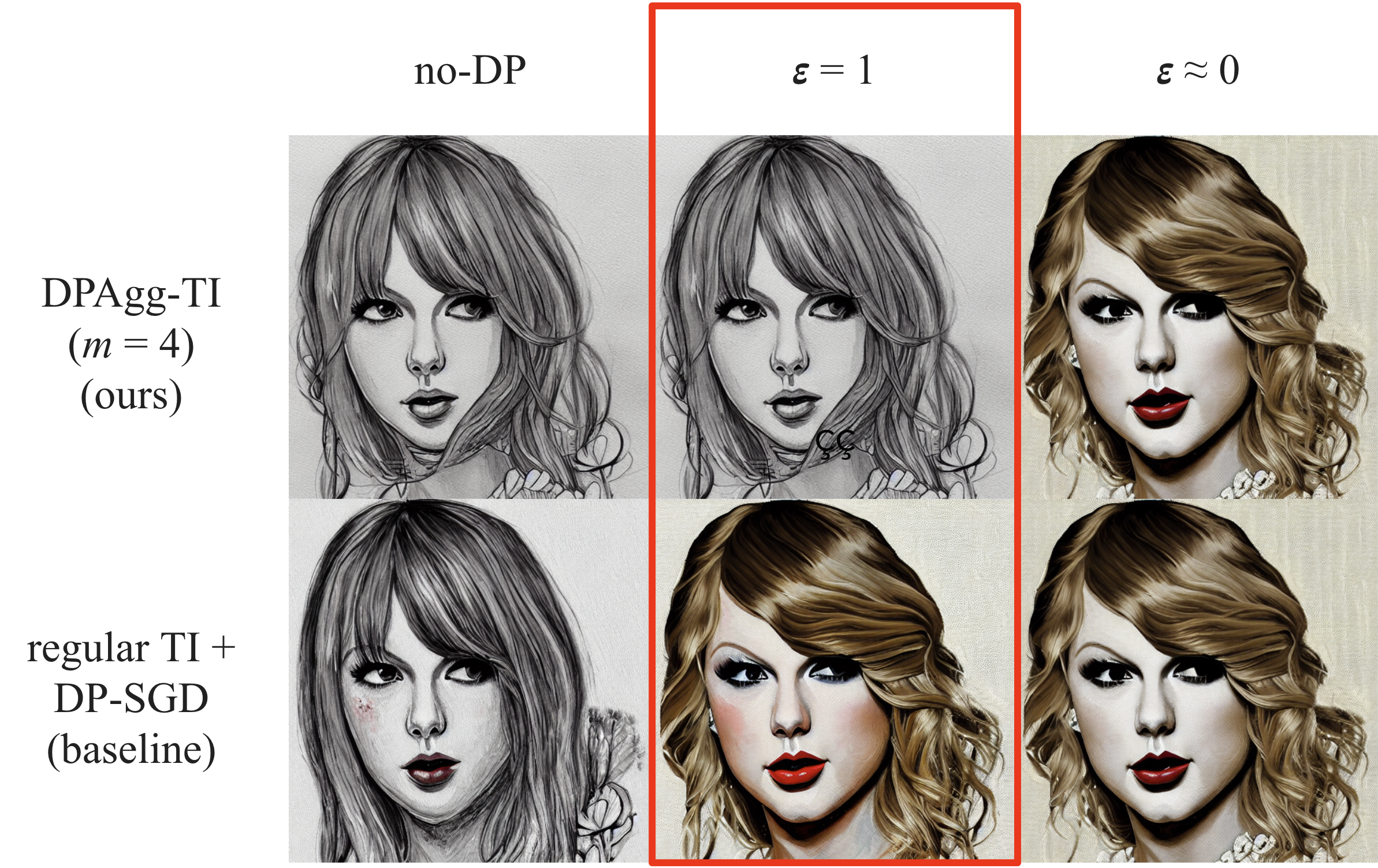}
    \caption{Comparing our approach to applying DP-SGD to regular TI using prompts ``an icon of a dragon in the style of the Paris 2024 Olympic Pictograms'' and ``a painting of Taylor Swift in the style of {\eve}'' respectively. Note that our method aggregates individual TI embeddings for each training image, whereas the baseline trains a single TI embedding over the entire dataset.}
    \label{fig:dp-sgd}
\end{figure}

\subsection{Ablation Study: Textual Inversion with DP-SGD}
A natural question that arises is how well our approach compares to the naive method of applying DP-SGD to regular TI training. We therefore integrated DP-SGD into the TI codebase using the Opacus library and trained similar embeddings on the {\eve} and Paris 2024 datasets. We found that in most cases, notably the {\eve} dataset, the amount of noise required for DP-SGD to achieve a reasonable value of $\varepsilon$ for DP is so high that the resulting embedding contains negligible information about the training dataset. In particular, the results for $\varepsilon = 1$ are almost indistinguishable to $\varepsilon \approx 0$, as shown in Figure \ref{fig:dp-sgd}. We believe that this is simply because DP-SGD is not designed to handle such small datasets in the order of 100 images. Additional results can be found in Appendix \ref{app:dpsgd}.

\section{Discussion}

\subsection{Copyright Protection Implications} \label{sec:copyright}

Our proposed mechanism can also be interpreted through the lens of \emph{copyright protection}. This connection is grounded in the framework of \emph{Near Access-Freeness (NAF)} \cite{vyas2023provable}, which evaluates whether a model’s outputs reveal undue influence from specific data points by comparing them to those from a safe model trained without access to the same data. 

Modern generative models typically produce outputs via randomized sampling. Leveraging this inherent randomness, Vyas et al. \cite{vyas2023provable} introduced NAF as a metric to quantify the similarity between a model's output and copyrighted content. The key idea is to compare the output distribution of a potentially infringing model to that of a \emph{safe} model---one trained without access to the target content. A canonical example is the \emph{leave-one-out-safe} model, trained on the full dataset excluding $x$. Since $\msf{safe}(x)$ lacks access to $x$, the probability that it generates content resembling $x$ is expected to be small; any such resemblance is considered fortuitous.

\begin{definition}[Near Access-Freeness \cite{vyas2023provable}]
Let $\mathcal{C}$ be a set of copyrighted samples and $\mathcal{W}$ a set of generative models.
Given a mapping $\msf{safe}: \mathcal{C} \rightarrow \mathcal{W}$ and a divergence measure $\Delta$,
we say a model $w \in \mathcal{W}$ is \emph{$k_y$-near access-free} (or $k_y$-NAF) on prompt $y \in \mathcal{Y}$
if for every $x \in \mathcal{C}$,
\[
\Delta\!\left( P_w(\cdot|y) \,\|\, P_{\msf{safe}(x)}(\cdot|y) \right) \leq k_y.
\]
\end{definition}

If $k_y = 0$, the model is indistinguishable from a safe model, meaning any resemblance to copyrighted material is by random chance. More generally, a small $k_y$ suggests the model is unlikely to generate outputs resembling $x$ with higher probability than a model that has never seen $x$.

NAF is closely related to concepts in DP. Depending on the divergence measure $\Delta$, NAF resembles different DP variants---for example, $\varepsilon$-DP when $\Delta = \Delta_{\mathrm{max}}$ \cite{dwork2006calibrating}, and $(1,\varepsilon)$-R\'enyi DP when $\Delta = \Delta_{\mathrm{KL}}$. Translating DP to generative models yields this definition:

\begin{definition}[Differentially Private Generation (DPG)]
Let $S$ and $S'$ be neighboring datasets. Denote by $P_S(\cdot|y)$ the distribution over outputs generated by a model trained on, or adapted from, $S$ with algorithm $\mathcal{A}$, where randomness includes both training and generation. The generation satisfies \emph{$\varepsilon$-Differentially Private Generation} ($\varepsilon$-DPG) if for every $y \in \mathcal{Y}$,
\[
\Delta\!\left( P_S(\cdot|y) \,\|\, P_{S'}(\cdot|y) \right) \leq \varepsilon.
\]
\end{definition}

Here, \emph{neighboring datasets} differ by a single data point (or privacy unit). If the training process is $\varepsilon$-DP, then the outputs naturally satisfy $\varepsilon$-DPG via the data processing inequality. One benefit of DPG is the flexibility to add noise during generation rather than training, potentially improving the utility–privacy tradeoff. However, there are notable distinctions: $\varepsilon$-DP offers protection under arbitrary post-processing and multiple outputs, whereas $\varepsilon$-DPG only guarantees privacy for single outputs. Also, under DP, the trained model can be released, but under DPG, only the outputs are safe to share. Elkin-Koren et al. \cite{elkin2024can} highlight further differences: NAF is \emph{one-sided}---comparing a model to a fixed safe reference---whereas DPG is \emph{symmetric}. This asymmetry in NAF can enable better utility. Additionally, NAF allows more flexibility in choosing the safe model, which can be exploited in algorithm design.

Since DPAgg-TI satisfies $\varepsilon$-DP, it also satisfies $\varepsilon$-NAF under the leave-one-out-safe model. Within the NAF framework, this means the adapted model behaves similarly to one that never saw the private images. Importantly, this guarantee is meaningful only \emph{within NAF}; it does not imply broader legal immunity or empirical indistinguishability from the original content. However, it allows us to argue that any close resemblance between outputs and private training data is no more likely than would be expected from a model with no access to that data.

Finally, the goal of DPAgg-TI is to adapt to the \emph{style} of private image sets, not their precise content.
This distinction matters:
pure style imitation (without reproducing protectable expression or ``substantive elements'') is often argued to be
non-infringing in many creative contexts, though the legal status is jurisdiction- and fact-dependent;
particularly in artistic and creative contexts. As discussed in Elkin-Koren et al. \cite{elkin2024can} and legal analyses such as Carlini et al. \cite{carlini2023extracting}, generating new content in the style of a work, without reproducing its substantive elements, is generally not considered copyright infringement. Therefore, the use of DPAgg-TI to learn and reproduce stylistic attributes does not contradict the spirit or intent of the NAF framework. Instead, it offers a promising direction for responsibly fine-tuning generative models on private or copyrighted sources while respecting both privacy and intellectual property boundaries.

\begin{remark}[Scope of Protection and Artist-Level Extension]
We provide \emph{record-level} DP: it limits leakage or reconstruction of any \emph{individual} private image, not an
artist's entire style. Consequently, it yields a corresponding NAF-style guarantee at the per-image level (under the chosen
safe reference). This interpretation should be understood strictly within NAF and does not constitute a general copyright
compliance claim. The DP guarantee continues to apply under targeted prompts: conditioning on detailed descriptions can
increase the likelihood of reproducing a specific private work by at most an $e^{\varepsilon}$ factor (up to $\delta$).
Artist-level (user-level) DP is conceptually possible by treating each artist as one unit and privately aggregating
artist-level embeddings (e.g., via a DP mean mechanism), but typically requires stronger noise and may reduce utility; we
leave a full exploration to future work.
\end{remark}

\subsection{Limitations}

DPAgg-TI is designed for the low-data, strong-privacy regime, where the number of private images is small ($n \approx 100$) and per-record protection with $\varepsilon < 5$ matters. For large datasets with moderate subsampling, DP-SGD on the full model may become more efficient and could provide better utility. We explicitly position our method for scenarios where DP-SGD is known to struggle: strong privacy guarantees with limited training data.

We acknowledge that in moderate privacy regimes ($5 \le \varepsilon \le 10$) with larger batch sizes and careful tuning, DP-SGD with parameter-efficient fine-tuning methods might perform better than our approach. However, in our experiments, applying DP-SGD to regular TI even at $\varepsilon = 20$ with carefully tuned hyperparameters (learning rate, batch size, scheduler) still required prohibitively high noise levels to satisfy the privacy accountant, preventing meaningful learning. Moreover, DP-SGD typically involves multiple training iterations, so the effective noise further accumulates due to composition across epochs, making convergence extremely difficult in the strong-privacy regime ($\varepsilon < 5$) with approximately 100 images.

\section{Conclusion}
We presented a differentially private adaptation method for diffusion models based on Textual Inversion, enabling privacy-preserving style transfer without the need for full model fine-tuning. By learning per-image embeddings and aggregating them with calibrated noise, our method, DPAgg-TI, achieves strong formal privacy guarantees while maintaining high output fidelity. Experiments on private artwork and Paris 2024 pictograms show that DPAgg-TI consistently outperforms DP-SGD, which fails to produce meaningful results under comparable privacy budgets. These results highlight the effectiveness of embedding-level adaptation as an efficient and scalable alternative to traditional gradient-based approaches, especially in low-data regimes. Unlike DP-SGD, which introduces significant computational overhead and utility degradation, DPAgg-TI is lightweight, modular, and compatible with existing diffusion backbones. Our findings suggest that embedding-centric approaches offer a promising direction for privacy-aware personalization, and motivate further research into cross-modal extensions, improved aggregation techniques, and integration with broader privacy-preserving frameworks.

\section*{Ethical Statement}

The use of images without owner consent raises significant ethical concerns, particularly regarding the exploitation of intellectual property. This work introduces a method for visual generative models to adapt to new styles and classes while ensuring privacy and copyright protection for data owners. By providing a framework for privacy-preserving adaptation, this technology aims to respect intellectual property and address ethical challenges in generative AI. While it does not eliminate the need for consent from data owners, we hope that it represents a step toward balancing innovation with ethical considerations in AI development. Beyond creative applications, the proposed method has broader potential uses, including synthetic data generation, privacy-preserving personalization, and fine-tuning diffusion models for private or domain-specific tasks. 

\section*{Acknowledgments}

We sincerely thank {\eve} for providing her artwork for use in this study. We are also grateful to Anwar Hithnawi and Varun Chandrasekaran for their insightful discussions and feedback, as well as to all participants in our user study. Sanmi Koyejo acknowledges support by NSF 2046795 and 2205329, IES R305C240046, the MacArthur Foundation, Stanford HAI, OpenAI, and Google. This work was supported in part by the National Research Foundation of Korea (NRF) grant funded by the Korea government (MSIT) (No. RS-2025-23525649).

\pagebreak

\input{custom.bbl}

\onecolumn%
\appendices

\section{Differentially Private Adaptation via Style Guidance}\label{app:Style Guidance}

\subsection{Background: Denoising Diffusion Implicit Models}

Denoising Diffusion Implicit Models (DDIM) sampling \cite{song2020denoising} uses the predicted noise $\epsilon_\theta(x_t, y, t)$ and a noise schedule represented by an array of scalars $\{\alpha_t\}_{t=1}^T$ to first predict a clean image $\hat{x}_0$, then makes a small step in the direction of $\hat{x}_0$ to obtain $x_{t-1}$. The reverse diffusion process for DDIM sampling can be formalized as
 \begin{equation}
     \hat{x}_0 = \frac{x_t - \sqrt{1 - \alpha_t}\epsilon_\theta(x_t, y, t)}{\sqrt{\alpha_t}}
\label{eq:ddim_predict}
 \end{equation}
\begin{equation}
     x_{t-1} = \sqrt{\alpha_{t-1}}\hat{x}_0 + \sqrt{1 - \alpha_{t-1}}\epsilon_\theta(x_t, y, t).
\label{eq:ddim}
\end{equation}

\subsection{Implementation}

We extend our approach to style guidance (SG) by leveraging the framework of Universal Guidance~\cite{bansal2024universal}.
Specifically, we focus on CLIP-based style guidance,
which optimizes the similarity between the CLIP embeddings of a target image and the generated image.

We encode each target image \(x^{(i)}\) as \(u^{(i)}\) via a CLIP image encoder,
then aggregate the embeddings \(u^{(1)}, \dots, u^{(n)}\) into \(u^*_{\text{DP}}\) 
using \eqref{eq:centroid_method} or \eqref{eq:centroid_subsample}, depending on whether subsampling is applied.
The aggregated embedding \(u^*_{\text{DP}}\) is then incorporated into the reverse diffusion process as a style guide.

Let \(x_c\) denote the target style image, \(x_t\) the noisy image at step \(t\), and \(\clip(\cdot)\) the CLIP image encoder.
The forward guidance process is defined as  
\begin{equation}
\hat{\epsilon}_\theta(x_t, y, t) = \epsilon_\theta(x_t, y, t) 
+ w\sqrt{1 - \alpha_t}\nabla_{x_t}\ell_{\cos}(\clip(x_t), \clip(\hat{x}_0)),
\label{eq:guidance}
\end{equation}
where \(w\) is a guidance weight and \(\ell_{\cos}\) is the negative cosine similarity loss. For a detailed description of Universal Guidance, including the backward guidance process and per-step self-recurrence, we refer the reader to the original paper. The reverse diffusion step replaces \(\epsilon_\theta(x_t, y, t)\) with \(\hat{\epsilon}_\theta(x_t, y, t)\),
generating an image \(x_0\) that aligns with the text conditioning \(y\) 
while incorporating the stylistic characteristics of \(x_c\).

To integrate differential privacy, we encode each target image \(x^{(i)}\) into \(u^{(i)} = \clip(x^{(i)})\) 
and aggregate these embeddings into \(u^*_{\text{DP}}\) using the centroid method.
The aggregated \(u^*_{\text{DP}}\) guides the reverse diffusion process: 
\begin{equation}
\hat{\epsilon}_\theta(x_t, y, t) = \epsilon_\theta(x_t, y, t) 
+ w\sqrt{1 - \alpha_t}\nabla_{x_t}\ell_{\cos}(u^*_{\text{DP}}, \clip(\hat{x}_0)).
\label{eq:guidance-dpagg}
\end{equation}

This ensures privacy-preserving style transfer while maintaining high stylistic fidelity.

\subsection{Style Transfer Results}

\begin{figure}[h]
\centering
\includegraphics[width=\columnwidth]{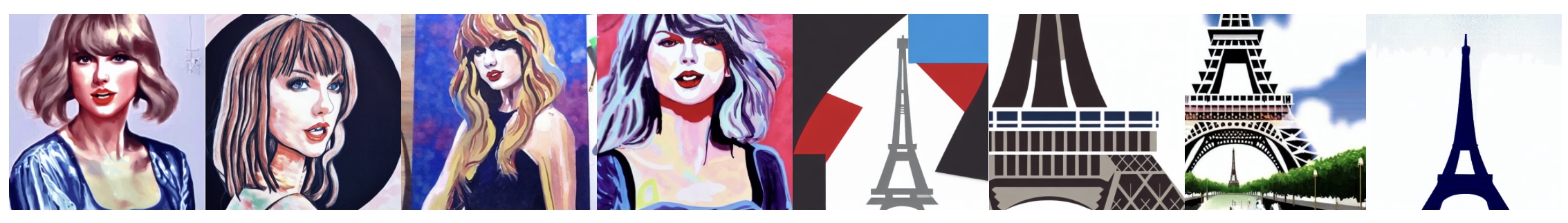}
\caption{Attempts of using universal guidance to generate drawings of Taylor Swift and icons of the Eiffel Tower in the styles of {\eve} and Paris 2024 Pictograms respectively. Here, we apply no subsampling or DP-noise.}
\label{fig:sg-results}
\end{figure}

We apply our SG-based approach to both datasets. 
While it provides privacy protection by obfuscating embedding details,
the resulting images captured only generalized stylistic elements 
and lack the detailed fidelity and coherence achieved with the TI-based method.
As shown in Figure~\ref{fig:sg-results}, this highlights the superiority of TI in balancing privacy and high-quality image generation.

The reduced effectiveness of SG for style transfer may stem from its sensitivity to hyperparameters 
such as the guidance weight $w$, leading to instability. 
Although Bansal et al. \cite{bansal2024universal} proposed remedies, namely backward guidance and per-step self-recurrence,
these proved insufficient for our application.
Additionally, the CLIP embeddings may not retain enough stylistic detail after the aggregation.

\subsection{Ablation}

\begin{figure*}[h]
\centering
\includegraphics[width=\textwidth]{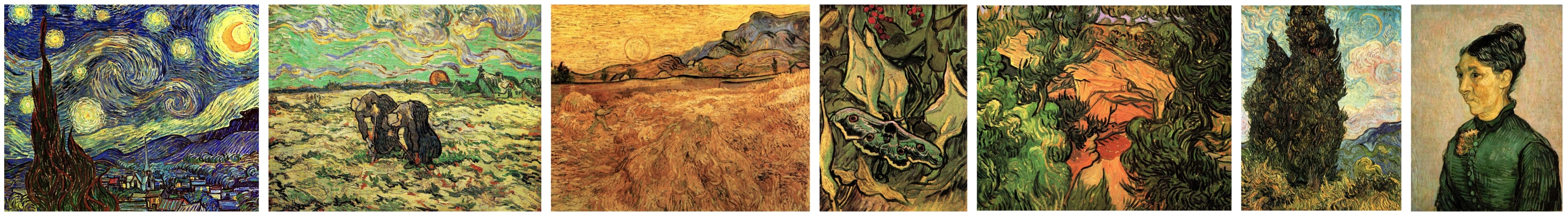}
\caption{Sample of paintings by Van Gogh used to generate style guidance embeddings.}
\label{fig:vgdata}
\end{figure*}

To better understand the limited effectiveness of style guidance in our experiments, despite its success in Bansal et al. \cite{bansal2024universal}, we applied our approach to a dataset of 143 paintings from Van Gogh’s Saint-Paul Asylum, Saint-Rémy collection\cite{vgdata} (Figure~\ref{fig:vgdata}). Unlike the {\eve} and Paris 2024 datasets, it is highly likely that Stable Diffusion has been trained on these images. Additionally, Bansal et al.~\cite{bansal2024universal} demonstrated successful adaptation towards the style of Van Gogh’s Starry Night as a single reference image, making this dataset a reasonable interpolation between their successful results and our more limited findings.

\begin{figure*}[h]
\centering
\includegraphics[width=\textwidth]{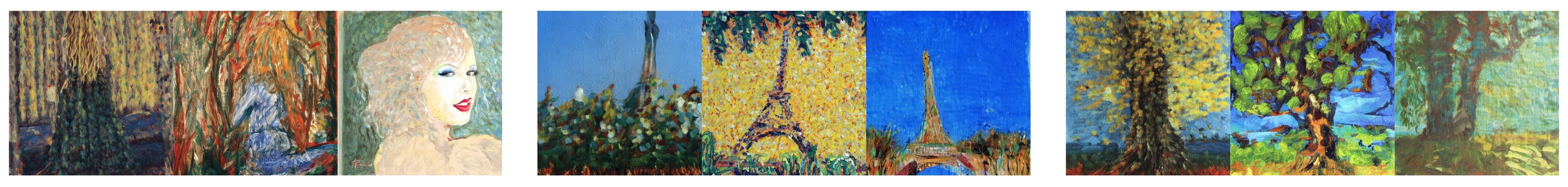}
\caption{Images generated by Stable Diffusion v1.5 with style guidance towards Van Gogh's \textit{Saint-Paul Asylum, Saint-Rémy} collection using prompts ``A painting of Taylor Swift (left) / the Eiffel Tower (center) / a tree (right)''.}
\label{fig:vgresults}
\end{figure*}

Without DP noise or subsampling, we obtained reasonable style transfer results, as shown in Figure~\ref{fig:vgresults}. This suggests that style guidance struggles when applied to previously unseen target styles, and that its effectiveness may depend on prior exposure within the pre-training data.


\section{Computational Cost Comparisons}

Direct comparisons of computational cost across methods are challenging 
due to differing training paradigms (per-image optimization vs.\ dataset-level training), optimization procedures,
and privacy accounting.
Nonetheless, to provide a concrete sense of scale,
we report representative costs measured using Stable Diffusion v1.5 on a single NVIDIA A100 GPU (Tables~\ref{tab:training_cost} and~\ref{tab:inference_cost}).
For each method, we tuned the number of optimization steps to reach its best utility under the target privacy budget.

\paragraph{Sequential vs.\ batched execution}
The per-image runtime reported in Table~\ref{tab:training_cost} corresponds to a sequential implementation
that optimizes each textual-inversion embedding independently.
In practice, we can optimize multiple embeddings jointly by batching several images at once 
(and optionally subsampling the private set per update),
which reduces the effective wall-clock cost and avoids the naive linear scaling 
implied by ``minutes per image $\times \; n$.'' 

\begin{table*}[ht]
\centering
\caption{
Training cost comparison across methods. Overhead from DP-SGD is relatively modest due to the low-dimensional embedding being optimized. N/A for SG means nothing is trained aside from the base model.
}
\begin{tabular}{lllll}
\hline
\textbf{Method} & \textbf{Steps} & \textbf{Batch Size} & \textbf{Time} & \textbf{Memory Usage} \\
\hline
TI (no DP)     & 10,000 (for 150 images) & 1 & 25 min     & 7 GB  \\

               &                         & 8 & 2.5 hours  & 20 GB \\

TI (DP-SGD)    & 30,000 (for 150 images) & 1 & 80 min     & 7 GB  \\

               &                         & 8 & 7 hours    & 20 GB \\

DPAgg-TI       & 2,000 per image         & N/A & $\sim$5 min/image & 7 GB \\

SG             & N/A                     & N/A & N/A        & N/A  \\
\hline
\end{tabular}
\label{tab:training_cost}
\end{table*}

\begin{table*}[ht]
\centering
\caption{
Inference cost comparison across methods. 
}
\begin{tabular}{lllll}
\hline
\textbf{Method} & \textbf{Steps} & \textbf{Batch Size} & \textbf{Time} & \textbf{Memory Usage} \\
\hline
TI (no DP, DP-SGD, DPAgg-TI) & 50  & 1   & 1–2 sec   & 4 GB  \\

                            & 100 & 1   & 1–2 min   & 58 GB \\

SG (no DP, DPAgg-SG)        & 500 & 1 & $\sim$30 min & 17 GB \\
\hline
\end{tabular}
\label{tab:inference_cost}
\end{table*}

\paragraph{When DP-SGD can be faster}
DP-SGD amortizes computation across the dataset and can be faster wall-clock-wise for larger $n$ 
and/or moderate privacy budgets.
In contrast, our method is designed for low-data personalization with strong per-record guarantees,
and it offers a practical advantage in dynamic settings:
it supports incremental updates to the private set (e.g. adding or removing images)
without retraining a large set of model parameters,
whereas DP-SGD-style training typically requires rerunning optimization to reflect such changes.
We view these approaches as complementary, targeting different operating regimes.

\section{Additional Style Transfer and Ablation Results}\label{app:dpsgd}

\begin{figure*}[h!]
\centering
\includegraphics[width=\textwidth]{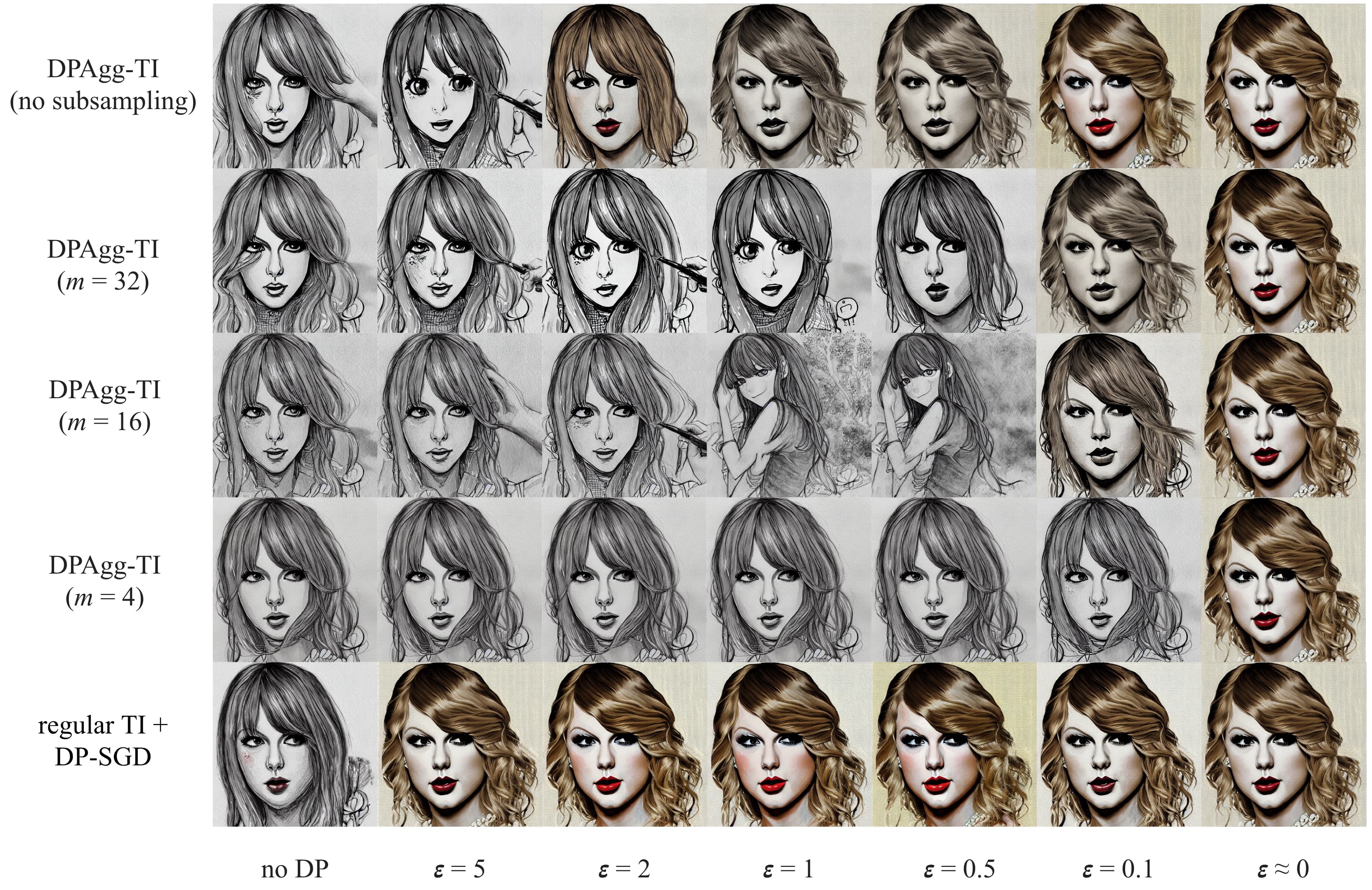}
\caption{Images generated by Stable Diffusion v1.5 using the prompt ``A painting of Taylor Swift in the style of $<${\eve}$>$'', 
with the embedding $<${\eve}$>$ trained using DPAgg-TI (with different subsample sizes $m$) and TI with DP-SGD using different values of $\varepsilon$.}
\label{fig:taylor-appendix}
\end{figure*}
\begin{figure*}[h!]
\centering
\includegraphics[width=\textwidth]{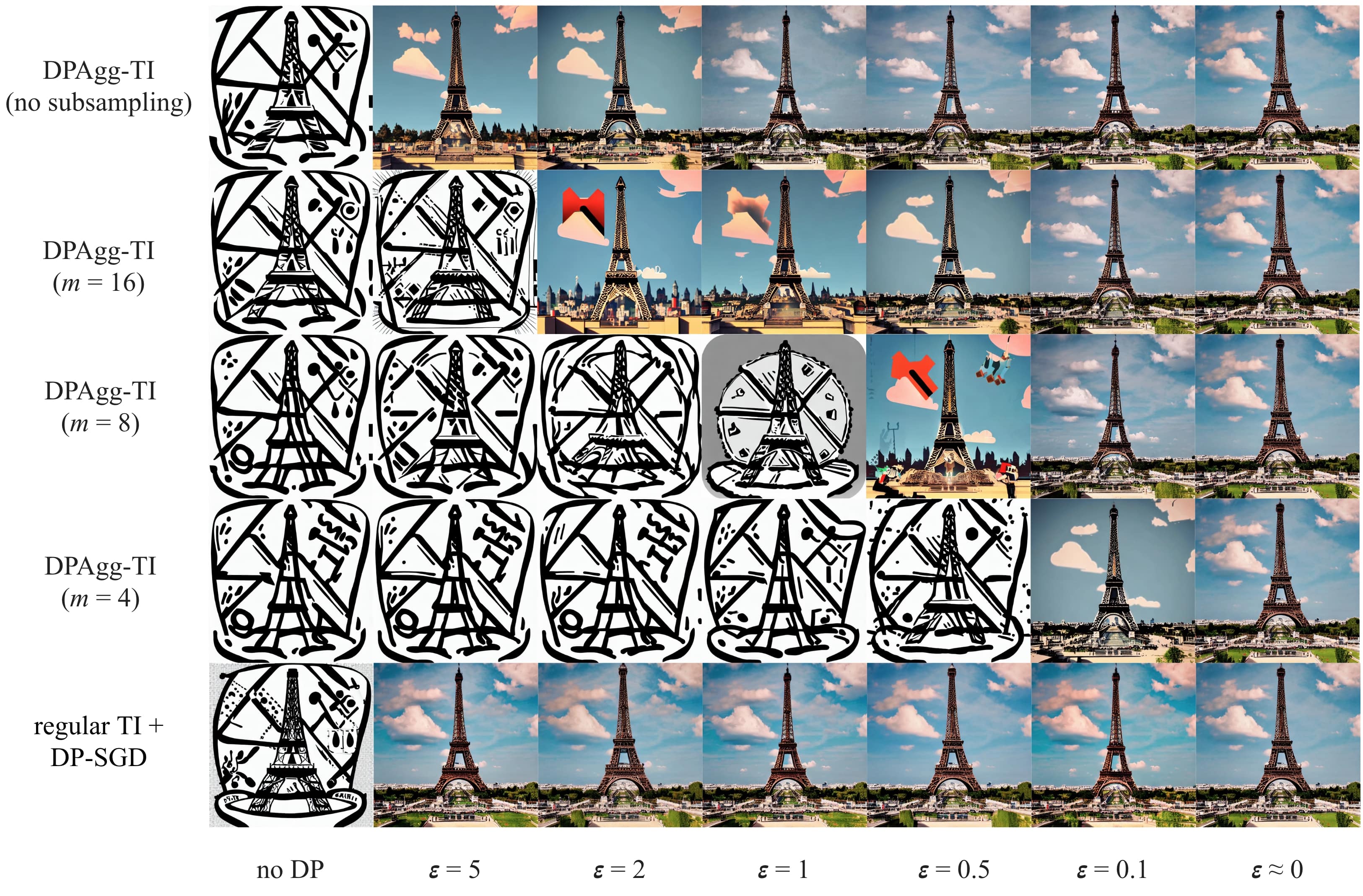}
\caption{Images generated by Stable Diffusion v1.5 using the prompt ``An icon of the Eiffel Tower in the style of $<${Paris 2024 Pictograms}$>$'', 
with the embedding $<${Paris 2024 Pictograms}$>$ trained using DPAgg-TI (with different subsample sizes $m$) and TI with DP-SGD using different values of $\varepsilon$.}
\label{fig:eiffel-appendix}
\end{figure*}
\begin{figure*}[h!]
\centering
\includegraphics[width=\textwidth]{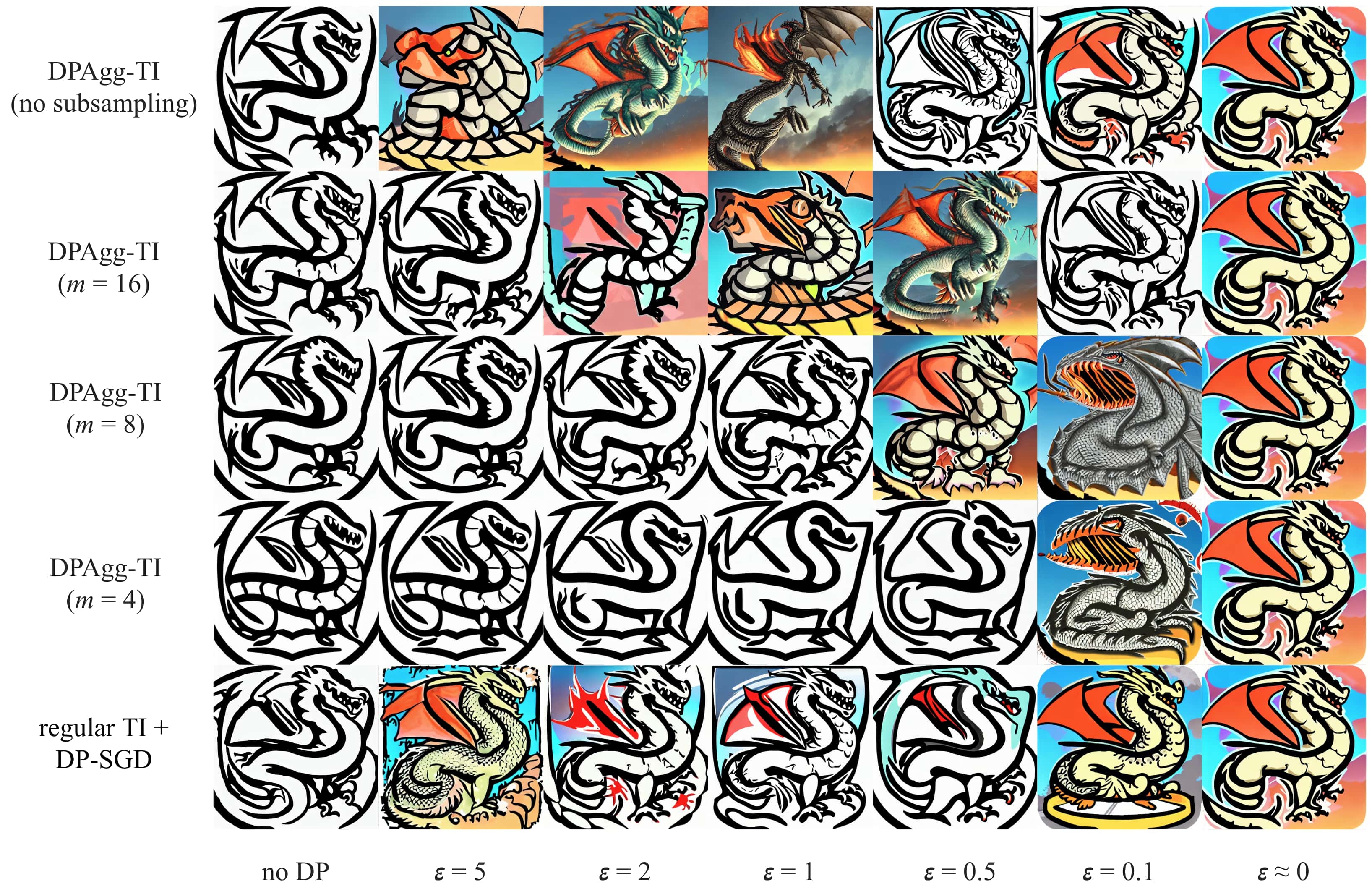}
\caption{Images generated by Stable Diffusion v1.5 using the prompt ``An icon of a dragon in the style of $<${Paris 2024 Pictograms}$>$'', 
with the embedding $<${Paris 2024 Pictograms}$>$ trained using DPAgg-TI (with different subsample sizes $m$) and TI with DP-SGD using different values of $\varepsilon$.}
\label{fig:dragon-appendix}
\end{figure*}

\end{document}